\pdfoutput=1

\documentclass[11pt]{article}

\usepackage[final]{acl}

\usepackage{times}
\usepackage{latexsym}

\usepackage[T1]{fontenc}

\usepackage[utf8]{inputenc}

\usepackage{microtype}

\usepackage{inconsolata}

\usepackage{graphicx}

\usepackage{amsmath,amsfonts,bm}

\def\eqref#1{equation~\ref{#1}}

\def\1{\bm{1}}

\DeclareMathAlphabet{\mathsfit}{\encodingdefault}{\sfdefault}{m}{sl}
\SetMathAlphabet{\mathsfit}{bold}{\encodingdefault}{\sfdefault}{bx}{n}

\usepackage{amsmath}
\usepackage{bbm}
\usepackage{amssymb}
\usepackage{mathtools}
\usepackage{amsthm}

\usepackage{algorithm}
\usepackage{algorithmic}
\usepackage{setspace}
\usepackage{comment}
\usepackage{booktabs}
\usepackage{hyperref}
\usepackage{icomma}
\usepackage{url}

\usepackage[nameinlink]{cleveref}

\usepackage{longtable}
\usepackage{xcolor}
\usepackage{colortbl}
\usepackage{multicol}
\usepackage{multirow}
\usepackage{makecell}
\usepackage{amsmath, amssymb}
\usepackage{mathtools}
\usepackage{enumitem}
\usepackage{subcaption}
\usepackage{float}
\usepackage{graphicx}
\usepackage{caption} 
\usepackage{upgreek}
\usepackage{seqsplit}
\usepackage{color,soul}
\usepackage{arydshln}
\usepackage{xspace}
\usepackage{adjustbox}
\usepackage{enumitem}
\usepackage{booktabs} 
\usepackage[most]{tcolorbox}

\definecolor{Red}{rgb}{0.6,0,0}
\definecolor{Blue}{rgb}{0,0,0.8}
\definecolor{Green}{rgb}{0,0.4,0.7}
\definecolor{mountainmeadow}{rgb}{0.19, 0.73, 0.56}
\definecolor{crimson}{rgb}{0.86, 0.08, 0.24}
\definecolor{darkblue}{rgb}{0.0, 0.0, 0.60}

\definecolor{gg}{HTML}{E0FEFE}
\definecolor{gray}{RGB}{236, 236, 236}

\hypersetup{
    colorlinks = true,
    citecolor = darkblue,
    linkcolor = darkblue,
    linktoc = all,
    urlcolor = darkblue,
}

\newcommand{\highlight}[1]{{\color{darkblue}{#1}}}

\newcommand{\ours}{PRInTS\xspace}

\title{PRInTS: Reward Modeling for Long-Horizon Information Seeking}

\author{
    Jaewoo Lee$^{1}$ \;
    Archiki Prasad$^{1}$ \;
    Justin Chih-Yao Chen$^{1}$ \\
    \textbf{Zaid Khan}$^{1}$ \;
    \textbf{Elias Stengel-Eskin}$^{2}$ \;
    \textbf{Mohit Bansal}$^{1}$ \\
    University of North Carolina at Chapel Hill$^{1}$, University of Texas at Austin$^{2}$\\
    \texttt{jwoolee@cs.unc.edu archiki@cs.unc.edu}
}

\begin{document}
\maketitle

\begin{abstract}
Information-seeking is a core capability for AI agents, 
requiring them to gather and reason over tool-generated information across long trajectories.
However, such multi-step information-seeking tasks remain challenging for agents backed by language models.
While process reward models (PRMs) can guide agents by ranking candidate steps at test-time, existing PRMs -- designed for short reasoning with binary judgment -- cannot capture richer dimensions of information-seeking steps, such as tool interactions and reasoning over tool outputs, nor handle the rapidly growing context in long-horizon tasks.
To address these limitations, we introduce \ours, a generative PRM trained with dual capabilities: (1) dense scoring based on the PRM's reasoning across multiple dimensions of step quality (e.g., interpretation of tool outputs, tool call informativeness) 
and (2) trajectory summarization that compresses the growing context while preserving essential information for step evaluation.
Extensive evaluations across FRAMES, GAIA (levels 1-3), and WebWalkerQA (easy-hard) benchmarks on multiple models reveal that best-of-$n$ sampling with \ours enhances information-seeking in open-source models as well as specialized agents, matching or surpassing frontier models with a much smaller backbone agent and outperforming other strong reward modeling baselines.\footnote{Code available at \href{https://github.com/G-JWLee/PRInTS}{https://github.com/G-JWLee/PRInTS}}
\end{abstract}

\section{Introduction}
\begin{figure*}[t]
    \centering
    \includegraphics[width=\linewidth]{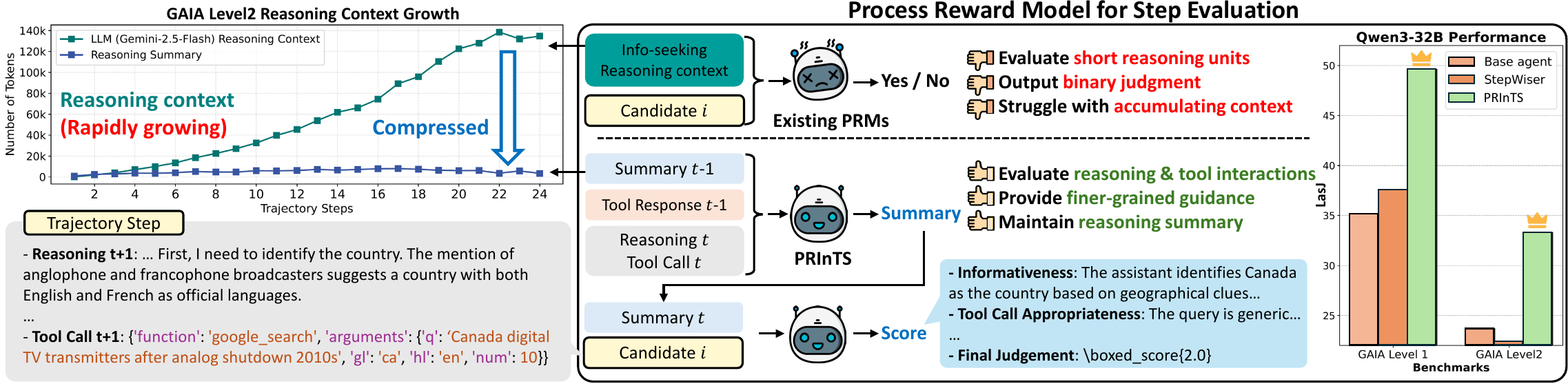}
    \par
    \caption{Comparison between existing PRMs and \ours. \textbf{Top}: Existing PRMs are limited for long-horizon information-seeking as they evaluate a short reasoning unit (e.g., one-to-two-sentence inferences) with coarse feedback, which cannot capture multi-faceted quality factors from tool interactions. They also struggle with rapidly accumulating reasoning context (left). \textbf{Bottom}: In contrast, \ours evaluates a complete trajectory step (reasoning + tool interactions), considers multiple trajectory step quality dimensions to produce dense scores for finer-grained guidance at each step, and maintains compact trajectory summaries that keep key information for the evaluation.
    }
    \label{fig:teaser}
\end{figure*}
A long-standing goal in artificial intelligence has been to develop agents that can answer novel queries by intelligently seeking information~\citep{bachman2016informationseekingagents,yuan202interactiveagent}, thereby enabling them to tackle challenging tasks in mathematics~\citep{Liu2025mmagent,Liu20025agenticmath}, software engineering~\citep{Yang2025swesmith,Pan2024swegym}, and research~\citep{Li2025websailor,Wu2025webdancer}. 
Large Language Models (LLMs) have shown promise as agents for such tasks when equipped with frameworks like ReAct~\citep{Yao2023react}, which interleaves LLM reasoning with external tool interactions.
However, long-horizon information-seeking tasks, which require agents to gather and synthesize information across multiple steps~\citep{Su2025bright,Shao2025reasonir}, remain challenging, even for recent LLMs with tool-use training, performing far below human-level~\citep{Mialon2024gaia,Wei2025browsecomp}.
While finetuning LLMs as information-seeking agents has shown promise~\citep{Li2025websailor,Wu2025webdancer}, it is limited to specific model families and is highly computationally demanding~\citep{Gao2025beyondtenturns}. An alternative way to boost a variety of agents is to build reward models (e.g., as done for math reasoning and instruction following~\citep{Wang2024mathshepherd,Zou2025reasonflux}). 
These models approximate the expected reward of a step or sequence of steps, enabling test-time scaling by ranking and selecting higher-quality actions or trajectories to successfully tackle long-horizon tasks.
Specifically, Process Reward Models (PRMs)~\citep{Zou2025reasonflux,Choudhury2025agentprm} offer a promising model-agnostic way of improving performance, scoring the quality of each of an agent's steps.

While past work has developed PRMs for tasks such as mathematics and logical reasoning, these methods are insufficient for long-horizon information-seeking tasks for two critical reasons.
(1) \textbf{Tool-Reasoning Evaluation Granularity}:
existing PRMs evaluate short reasoning units in isolation, typically one- to two-sentence logical or mathematical inferences~\citep{Xiong2025stepwiser,Zhao2025genprm}, providing binary judgments based on logical/math validity.
In contrast, long-horizon information-seeking requires jointly evaluating a complete trajectory step, which encompasses a reasoning step combined with tool interactions (e.g., web search, web browsing, code execution).
Moreover, step quality depends on multiple factors (e.g., interpretation of tool outputs, tool call informativeness, next-action plan) that coarse feedback cannot capture, requiring more granular guidance to effectively steer agents toward good trajectories.
(2) \textbf{Context Accumulation}: existing PRMs cannot manage the ever-growing reasoning context arising over multiple trajectory steps.
As illustrated in~\Cref{fig:teaser} \highlight{(Top-left)}, the information-seeking trajectory -- interleaving reasoning steps, tool calls, and tool call outputs -- grows rapidly as tool responses at each step introduce lengthy content.
Since models struggle to process such long, accumulated contexts~\citep{tang2025longrm,yen2025lostmaze,babilong}, often resulting in noisy evaluations, it necessitates compressing trajectories into compact forms rather than processing the full history.

Our work aims to fill these gaps by introducing 
\textbf{P}rocess \textbf{R}eward via \textbf{In}formation gain scoring and \textbf{T}rajectory \textbf{S}ummarization (\ours), a novel generative PRM for long-horizon information-seeking tasks.
\ours is a unified model jointly trained with two key abilities to address both the need for fine-grained guidance and the challenge of context accumulation. These two abilities are learned jointly within the same PRM.
First, \ours acts as a scorer that evaluates candidate next trajectory steps by generating Chain-of-Thought~\citep{Wei2025cot} analyses across multiple quality dimensions and outputting dense scores derived from this generative reasoning, as illustrated in (\Cref{fig:teaser} \highlight{(Bottom)}). 
Crucially, we frame step evaluation as information gain estimation that quantifies how much each trajectory step increases the probability of reaching the correct answer.
This formulation enables training via reinforcement learning with information gain estimation and preference prediction objectives, providing richer reward signals that account for the multi-faceted quality of trajectory steps. 
At test-time, \ours evaluates $n$ candidate next steps, selecting the step expected to yield the greatest information gain.
Second, \ours simultaneously functions as a summarizer, recursively generating and updating a compact trajectory summary at each step.
\ours compresses the query, previous summary, latest tool response, and current step into an updated summary that captures essential findings and plans up to the current timestep (\Cref{fig:teaser} \highlight{(Bottom)}). 
This keeps input length bounded, as shown in~\Cref{fig:teaser} \highlight{(Top-left)}, while preserving information for subsequent evaluation.

To equip \ours with these dual capabilities, we first design preference and summary data that can produce supervision signals needed to train the scoring and summarization components.
Specifically, our annotation pipeline (\Cref{fig:data_annotation})
uses Monte Carlo rollouts~\citep{Wang2024mathshepherd,Setlur2025pav} to estimate information gain scores and construct preference trajectory step pairs, and generates compact trajectory summaries for each step.
Next, we use this annotated data to train \ours to score steps via reinforcement learning with two complementary rewards: 
(1) a \textbf{Score Reward} that teaches the model to analyze the trajectory step quality and estimate the step's information gain score, and (2) a \textbf{Comparison Reward} that teaches the model to assign higher scores to preferred trajectory steps by learning from pairwise preferences. 
These rewards enable the model to capture the multi-faceted quality of the trajectory step and perform dense step-level evaluation. 
We jointly train \ours via supervised fine-tuning for summarization that recursively updates the trajectory summary based on the previous summary and the most recent reasoning context at each step, directly addressing the context accumulation challenge while preserving key information needed for step-level evaluation.
Together, this pipeline enables \ours to serve as a unified PRM capable of both managing long, noisy trajectories and providing fine-grained test-time guidance.

We validate our approach across three distinct LLMs used as information-seeking agents: Qwen3-32B~\citep{qwen3}, Tongyi DeepResearch-30B-A3B~\citep{tongyi2025deepresearch} -- a specialized information-seeking agent -- and Gemini-2.5-Flash~\citep{gemini}, evaluated on three long-horizon information-seeking benchmarks: FRAMES, GAIA, and WebWalkerQA.
The experimental results show that \ours, a 4B PRM, consistently provides test-time gains across diverse agents -- Qwen3-32B by 9.3\%, DeepResearch-30B-A3B by 3.9\%, and Gemini-2.5-Flash by 4.0\% absolute average accuracy -- without fine-tuning the underlying models. 
Unlike existing PRMs, which obtain diminished and inconsistent gains as agents become stronger, \ours continues to deliver substantial improvements.
Notably, on GAIA (levels 1-3), \ours raises DeepResearch-30B-A3B from 61.9\% to 64.4\% in our implementation, enabling the 30B agent augmented with the 4B PRM to outperform Deepseek-V3.1-671B (63.1\%) — which is 20 times larger — while also closing the gap to a strong proprietary system like OpenAI DeepResearch (67.4\%).
Furthermore, our ablation studies reveal that providing compressed summaries outperforms using raw trajectories as input context, showing that context management is essential for accurate step-level evaluation in long-horizon tasks.
Overall, our approach enhances information-seeking abilities of pretrained open-source models as well as specialized agents. showing strong generalizability.

\section{Related Work}
\label{sec:related_work}

\paragraph{Large Language Models (LLM) as Agents.}
LLMs have been increasingly adopted as agents through frameworks such as ReAct~\citep{Yao2023react}, which interleaves reasoning and tool use to solve complex tasks~\citep{Deng2025atomsearcher,Wu2025mindmap}.
To facilitate effective information-seeking behaviors, recent studies, such as WebSailor~\citep{Li2025websailor}, WebShaper~\citep{Tao2025webshaper}, and DeepResearch~\citep{tongyi2025deepresearch}, improve the intrinsic quality of information-seeking trajectories by training models on synthetic data to reduce search space uncertainty.
However, such methods share key limitations: they require (1) substantial supervision~\citep{2025mirothinker,tongyi2025deepresearch} and (2) access to model weights, which pose challenges for generalization whenever the underlying model changes.
We empirically show that \ours enhances information-seeking capabilities of agents via test-time guidance with strong generalizability, offering an orthogonal yet mutually beneficial direction to agent fine-tuning.

\paragraph{Reward Models for Reasoning.}
Outcome Reward Models (ORMs) predict the correctness of complete reasoning trajectories~\citep{Kim2024prospector, Pan2024swegym} but cannot provide finer-grained, step-wise guidance over partial trajectories.
Process Reward Models (PRMs) address this limitation by evaluating individual steps~\citep{Ton2024infogainprm,He2025generativeprm,chen-etal-2025-magicore}. 
Recent advancements cast PRMs 
as generative judges~\citep{Wang2024selftaughtevaluators,Whitehouse2025j1,He2025generativeprm} that generate justifications for step scores and have achieved strong performance in mathematics~\citep{Zhao2025genprm,Xiong2025stepwiser,Wang2024mathshepherd}, finance~\citep{Zhou2025finprm}, and agentic tasks~\citep{Chae2025webshepherd, Choudhury2025agentprm}.
In contrast to these existing PRM approaches that rank the validity of short reasoning snippets or struggle with managing growing contexts, \ours is equipped with jointly evaluating reasoning with tool interactions, planning for subsequent actions across multiple dimensions of ``information gain'' in tandem with a compact trajectory summarization mechanism, inspired by reasoning context compression approaches~\citep{Wu2025mindmap, kang2025acon,Ye2025agentfold}.

\begin{figure*}[t]
    \centering
    \includegraphics[width=\linewidth]{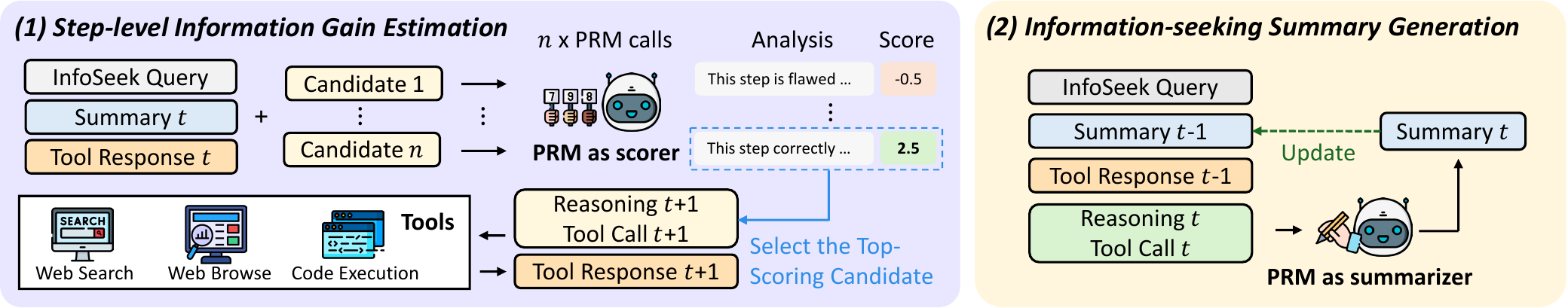}
    \par
    \caption{\textbf{Overview of \ours. \textbf{Left}}: \ours functions as a scorer, evaluating agent's multiple candidate next trajectory steps based on the summarized context and current tool response. It generates an analysis and a dense score for each candidate, selecting the top-scoring one to guide the agent's information-seeking. \textbf{Right}: \ours acts as a summarizer, recursively updating a compact information-seeking trajectory summary to preserve key information for subsequent score evaluation.}
    \label{fig:overview}
\end{figure*}
\section{Process Reward via Information Gain Score and Trajectory Summarization}
\label{headings}
We start by introducing the framework that quantifies and annotates the quality of each trajectory step -- a reasoning step combined with a tool call -- followed by reinforcement learning that uses these annotations to train \ours as a scorer (\Cref{sec:sub:prm_score}). 
Next, we describe our approach for generating compact summaries of long interaction trajectories, and explain how these summaries are used to train the same \ours as a summarizer (\Cref{sec:sub:prm_summary}). The overall design of \ours is illustrated in~\Cref{fig:overview}.

\subsection{Preliminaries}\label{sec:sub:prelim}
To tackle long-horizon information-seeking problems, we adopt the agentic ReAct~\citep{Yao2023react} paradigm, where a Large Language Model (LLM) acts as an agent that interleaves reasoning with tool-based action toward the goal of answering query $q$~\citep{Tao2025webshaper,tongyi2025deepresearch}. 
At each timestep $t$, the agent may generate intermediate reasoning $s_t$ based on the current context and then predict the subsequent action $a_t$, i.e., calls to external tools, such as web search, web browsing, and code execution, to acquire new information.  
The resulting tool response $o_t$ is observed and added to the context, which informs the agent's reasoning at timestep $t+1$.
\Cref{fig:overview} \highlight{(Left)} visually shows this tool interaction process: 
$s_t$, $a_t$, $o_t$ correspond to reasoning, tool call, tool response at timestep $t$.
This process repeats until timestep $T$, when the agent submits its answer $o_{T}$, which is successful if $o_T$ matches the ground-truth answer $a^{*}$. 
The accumulated reasoning context up to timestep $t$, termed the information-seeking trajectory, is defined as:
\begin{equation}
H_{t}=\left(s_{1},\;a_{1},\;o_{1},\;s_{2},\;a_{2},\;o_{2},\;\ldots,\;s_{t},\;a_{t},\;o_{t}\right)
    \label{eq:trajectory}
\end{equation}
Specifically, the agent $\pi$ generates the next reasoning step and tool call conditioned on the query and information-seeking trajectory, i.e., $s_{t}, a_{t} \sim\pi\left(\cdot|q,H_{t-1}\right)$.
Then the tool call is executed to get the tool response $o_t$. 
This interleaving of reasoning and action has shown success in long-horizon information-seeking tasks~\citep{Li2025websailor,Gao2025beyondtenturns,tongyi2025deepresearch}, and thus we adopt this setting.
However, applying PRMs here faces two key challenges: trajectory steps contain substantially richer content than traditional steps, requiring multi-dimensional evaluation beyond simple correctness, and rapidly growing context $H_t$ introduces noise that complicates evaluation.
Thus, we introduce our data annotation and train pipeline, which equips \ours with two core capabilities: (1) dense step-level scoring for fine-grained guidance, and (2) trajectory summarization for effective step-level evaluation under context accumulation.

\subsection{Step-level Information Gain Estimation}\label{sec:sub:prm_score}
\paragraph{Information Gain Score.}
\begin{figure*}[t]
    \centering
    \includegraphics[width=\linewidth]{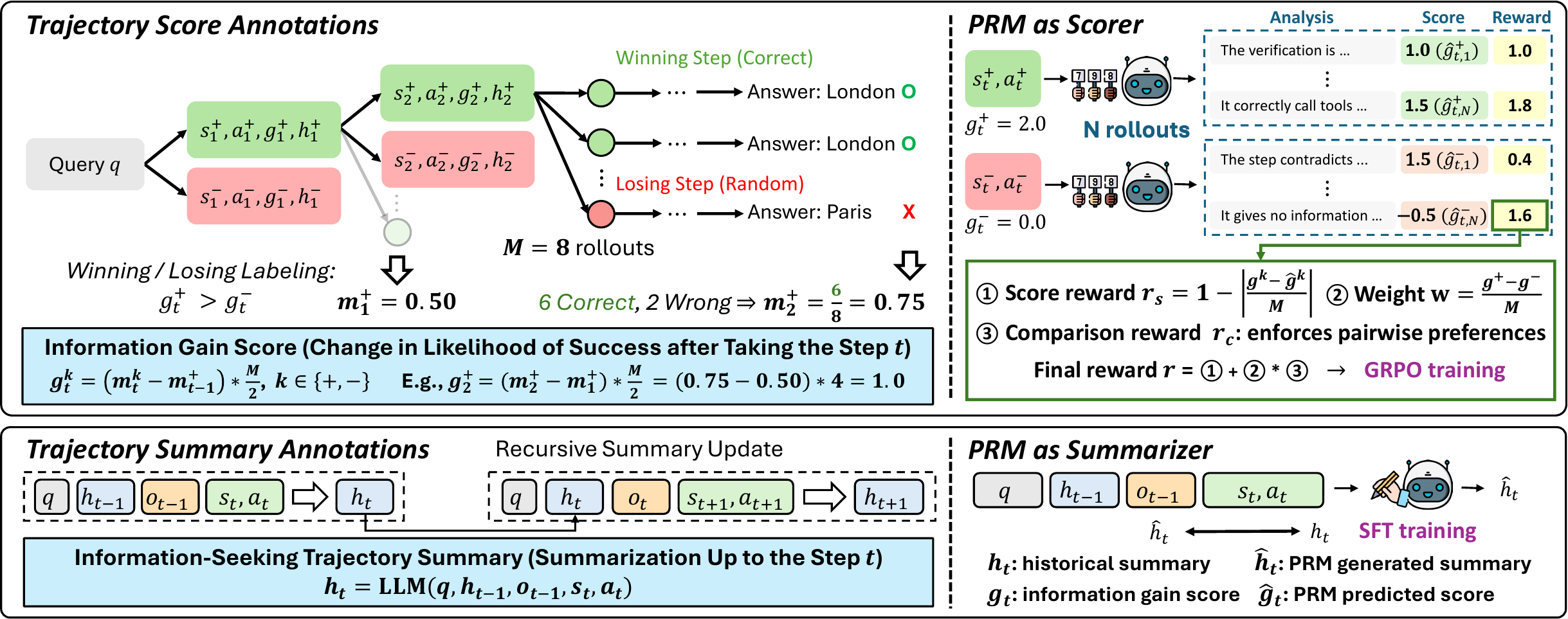}
    \par
    \caption{\textbf{\ours: data annotation and training pipeline.}  \textbf{Top}: For each trajectory step, we estimate the information gain score via Monte Carlo rollouts as the change in mean answer accuracy before and after the step. Then we construct winning-losing step pairs based on these scores (left). Preference pair examples are shown in~\Cref{fig:win_lose_examples}. Then we train \ours as a scorer via GRPO on these pairs (right). The final reward combines a score reward for accurate prediction, a comparison reward for pairwise preference learning, and an adaptive weight to mitigate noisy annotations.
    \textbf{Bottom}: Each step is annotated with a compact, recursively updated trajectory summary capturing essential findings and plans up to the step (left). The same PRM is jointly trained as a summarizer via SFT on this summary data (right). 
    }
    \label{fig:data_annotation}
\end{figure*}
To train a Process Reward Model (PRM) for long-horizon information seeking, we measure how much each reasoning step and tool call contributes towards reaching the correct answer.
To this end, we define information gain of the current step as the change in expected likelihood of arriving at the correct answer before and after taking the current step~\citep{Rao2018askgoodquestion,Prasad2023receval,Wang2024mathshepherd}.
This local evaluation quantifies the marginal improvement in task success contributed by the current step.
Specifically, for a reasoning step and tool call $(s_t, a_t)$ preceded by information-seeking trajectory prefix, $H_{t-1}$, we use Monte Carlo estimation~\citep{Wang2024mathshepherd,Xiong2025stepwiser,Setlur2025pav}
by executing $M$ rollouts until their final answers are produced and compute the mean accuracy:
\begin{equation}
    m_{t}=\frac{\sum_{j=1}^{M} \mathbbm{1}(o_{T_j}^{(j)} = a^*)}{M},
    \label{eq:mean_acc}
\end{equation}
where $o_{T_j}^{(j)}\sim\pi(\cdot|q,H_{t-1},s_t,a_t)$ is the final answer from rollout $j$, which terminates at timestep $T_{j}$.
The information gain score $g_t$ is then computed as: 
\begin{equation}
    g_{t} = \left(m_{t}-m_{t-1}\right)\times M/2,
    \label{eq:information_gain}
\end{equation}
which quantifies how much $(s_t, a_t)$ contributes to the successful completion of the task.
We scale by $M/2$ to map the scores into the interval $[-M/2, M/2]$ with discrete increments of 0.5.
This normalization provides a more intuitive understanding of the relative quality differences across steps.
A positive $g_t$ indicates that the current step $(s_t, a_t)$ increases the probability of reaching the correct answer -- for example, through logically coherent reasoning or a tool call that resolves uncertainties -- whereas a $g_t$ lower than zero indicates that the current step reduces the probability, e.g., by making unverified assumptions or invoking an irrelevant tool call.

\paragraph{Trajectory Score Annotations.} 
While prior approaches annotate individual trajectory in isolation~\citep{Xiong2025stepwiser,Wang2024mathshepherd} or use imitation learning on step-level scores~\citep{Wang2024mathshepherd}, 
\citet{Whitehouse2025j1} shows that pairwise preference learning builds robust judges. 
We extend this by automatically constructing preference pairs grounded in information gain scores, then training \ours with complementary objectives -- a score reward for information gain estimation and a comparison reward for preference prediction -- as illustrated in~\Cref{fig:data_annotation} \highlight{(Top)}.

During the $M$ rollouts for $g_t$, the LLM generates a set of $M$ unique next trajectory steps $\{s^{(j)}_{t+1},a^{(j)}_{t+1}\}_{j=1}^{M}$ and the corresponding final answer predictions $\{o_{T_j}^{(j)}\sim\pi\left(\cdot|q,H_t,s^{(j)}_{t+1},a^{(j)}_{t+1}\right)\}_{j=1}^{M}$. 
As shown in~\Cref{fig:data_annotation} \highlight{(Top-left)}, to construct a candidate preference pair, we first select the step that yields a successful trajectory, assuming this step has the highest potential to be effective among the $M$ rollouts.
A second trajectory step is then randomly sampled from the remaining steps, serving as a contrasting, less effective (i.e., incorrect or inefficient) alternative.

Next, we annotate the information gain scores of this candidate preference pair by treating each trajectory step as a new starting point and running $M$ rollouts to estimate their respective mean accuracies and information gain scores.
After annotation, we reassign preference labels based on these scores: the step with the higher score becomes the winning sample $(s^+_{t+1},a^+_{t+1})$, while the other becomes the losing sample $(s^-_{t+1},a^-_{t+1})$.
The winning step then serves as the starting point for generating the next pair at step $t+2$.
This contrastive labeling ensures that \ours learns relative preferences between trajectory steps grounded in empirical improvements. 

\paragraph{Training the PRM as a Scorer.}
The core function of \ours is to assess trajectory step quality and assign higher scores to steps expected to yield greater information gain.
To this end, we train \ours to evaluate step quality by predicting information gain scores. 
Given a query $q$, trajectory summary $h_{t-1}$ (introduce in~\Cref{sec:sub:prm_summary} below), latest tool response $o_{t-1}$, and trajectory step $(s_t, a_t)$, \ours generates a Chain-of-Thought analysis and outputs a scalar score $\hat{g}_t = f_I(q, h_{t-1},o_{t-1},s_t,a_t)$, where $f_{I}$ denotes \ours that works as an information gain scorer function.
We train this scoring capability using GRPO ~\citep{Shao2024grpo} with the following rewards: (1) a score reward ($r_s$) that targets minimizing the discrepancy between the predicted score $(\hat{g}_{t})$ and the ground-truth score $(g_{t})$, and (2) a comparison reward ($r_c$) that enforces pairwise preferences derived from annotated pairs:
\begin{equation}
    r^{k}_{s}\!=\!1\!-\!\left|\frac{g^{k}\!-\!\hat{g}^{k}}{M}\right|, r^{k}_{c}\!=\!\sum_{j=1}^{N}\!\frac{y^k\cdot\mathrm{sgn}(\hat{g}^{k}\!-\!\hat{g}^{\bar{k}}_{j})}{N},
    \label{eq:reward_definition}
\end{equation}
where $\operatorname{sgn}(x)\!=\!\begin{cases}+1,& x\ge 0\\-1,& x<0\end{cases}$
, $k\!\in\!\{+,\;-\}$ indicates the winning or losing sample, $y^{+}=1$ and $y^{-}=-1$ for comparison direction, $\bar{k}$ denotes its counterpart, $N$ is the number of rollouts, and $\hat{g}_{j}^{\bar{k}}$ is the predicted score of $j$-th rollout from the counterpart. For simplicity, we omit rollout- and step-level indices, which do not affect the underlying formulation of the scores. 
The comparison reward ensures \ours learns to distinguish better from worse reasoning paths, while the score reward provides fine-grained feedback on estimation accuracy.
Finally, we combine the two rewards into a single scalar per rollout with an adaptive weight:
\begin{equation}
    r^{k}=r^{k}_{s} + w*r^{k}_{c},\;\;w=\frac{g^{+}-g^{-}}{M},
    \label{eq:final_reward}
\end{equation}
where $w$ is the comparison reward weight set for each pair.
This adaptive weighting scheme addresses noise in the automatically annotated pairs. 
Pairs with large score margins $(g^{+}-g^{-})$ are more reliably ranked and receive higher weights, while pairs with small margins receive lower weights, as they may reflect annotation noise rather than true preference~\citep{Prasad2024selfconsistency}. 
The combined reward thus encourages \ours to estimate absolute scores accurately and to learn robust preferences in information-seeking trajectories. The illustration of reward computation is shown in~\Cref{fig:reward_compute}.

\subsection{Info-seeking Summary Generation}\label{sec:sub:prm_summary}
\paragraph{Trajectory Summary Annotations.}
Another core challenge in building a PRM for information-seeking agents is the rapidly growing context (\Cref{fig:teaser} \highlight{Top-left}) of lengthy and noisy reasoning and tool interactions.
This context explosion hinders PRMs from efficient processing and results in noise and distraction in quality evaluation.
To address this, we extract a concise summary of the information-seeking trajectory of each trajectory step $(s_{t},a_{t})$.
This summary, $h_t$, captures the essential findings and plan development up to timestep $t$.
As illustrated in~\Cref{fig:data_annotation} \highlight{(Bottom-left)}, each summary is recursively updated and generated by an LLM, incorporating the previous summary $h_{t-1}$, latest tool response $o_{t-1}$, and current trajectory step $(s_{t}, a_{t})$ (i.e., $h_t=\text{LLM}(q,h_{t-1},o_{t-1},s_{t},a_{t})$).
This recursive formulation ensures that $h_{t}$ maintains a compressed form of the entire trajectory $H_t$.

\paragraph{Training the PRM as a Summarizer.}
To enable efficient processing of the context during the score estimation, \ours learns to generate a concise summary $\hat{h}_t = f_S(q, h_{t-1},o_{t-1},s_t,a_t)$ that retains essential information, where $f_S$ is \ours that works as a summarization function.
We use supervised fine-tuning (SFT) on the annotated summaries $h_t$, effectively teaching the model to compress the context by imitating the annotations.

\begin{table*}[t]
    \small
    \centering
        \renewcommand{\arraystretch}{1.05}
        \setlength{\tabcolsep}{10pt}
        \begin{tabular}{l c c c c c c c c}
             \toprule
             {\multirow{2}{*}{\textbf{Method}}} & {\multirow{2}{*}{\textbf{FRAMES}}} & \multicolumn{3}{c}{\textbf{GAIA}} & \multicolumn{3}{c}{\textbf{WebWalkerQA}} & {\multirow{2}{*}{\textbf{Avg.}}}\\
             \cmidrule(l{0.5em}r{0.5em}){3-5} \cmidrule(l{0.5em}r{0.5em}){6-8}
              & & {\textbf{Level 1}} & {\textbf{Level 2}} & {\textbf{Level 3}} & {\textbf{Easy}} & {\textbf{Medium}} & {\textbf{Hard}} & \\
             \midrule
             Base agent &
             {\small 49.3} & {\small 35.1} & {\small 23.7} & {\small 11.1} & {\small 30.1} & {\small 26.9} & {\small 30.3} & {\small 29.5} \\
             \cmidrule{0-8}
             Confidence &
             {\small 55.7} & {\small 36.8} & {\small 24.4} & {\small \underline{16.7}} & {\small 31.7} & {\small 31.3} & {\small 32.9} & {\small \underline{32.8}} \\
              Relevance &
             {\small \underline{56.3}} & {\small 34.2} & {\small 20.5} & {\small 8.3} & {\small \underline{33.3}} & {\small 29.5} & {\small 32.5} & {\small 30.7} \\
             Verbal-progress &
             {\small 45.0} & {\small 35.9} & {\small 21.2} & {\small 13.9} & {\small 27.6} & {\small 30.2} & {\small 34.2} & {\small 29.7} \\
              \cmidrule{0-8}
             GenPRM-7B &
             {\small 50.0} & {\small 32.5} & {\small \underline{25.7}} & {\small \underline{16.7}} & {\small \underline{33.3}} & {\small \underline{32.8}} & {\small \underline{34.6}} & {\small 32.2} \\
             Web-Shepherd-8B &
             {\small 49.0} & {\small \underline{38.5}} & {\small 23.7} & {\small 5.5} & {\small 28.5} & {\small 31.8} & {\small 33.3} & {\small 30.0} \\
             StepWiser &
             {\small 51.3} & {\small 37.6} & {\small 22.4} & {\small 8.3} & {\small 31.7} & {\small 31.8} & {\small 33.8} & {\small 31.0} \\
              \cmidrule{0-8}
             \cellcolor{gg} \ours &
             \cellcolor{gg}{\small \textbf{58.7}} & \cellcolor{gg}{\small \textbf{49.6}} & \cellcolor{gg}{\small \textbf{33.3}} & \cellcolor{gg}{\small \textbf{19.4}} & \cellcolor{gg}{\small \textbf{39.8}} & \cellcolor{gg}{\small \textbf{33.3}} & \cellcolor{gg}{\small \textbf{37.3}} & \cellcolor{gg}{\small \textbf{38.8}} \\
             \bottomrule
        \end{tabular}
    \caption{Comparison of step quality evaluation methods on Qwen3-32B across information-seeking benchmarks. We adopt the LLM-as-Judge (LasJ) metric and report Avg@3. The best and the second best results are in \textbf{bold} and \underline{underline}, respectively. \ours delivers consistent gains all benchmarks, whereas the second-best baseline varies.
    }
    \label{tab:qwen3_eval}
\end{table*}
\begin{table*}[t]
    \small
    \centering
        \renewcommand{\arraystretch}{1.05}
        \setlength{\tabcolsep}{10pt}
        \begin{tabular}{l c c c c c c c c}
             \toprule
             {\multirow{2}{*}{\textbf{Method}}} & {\multirow{2}{*}{\textbf{FRAMES}}} & \multicolumn{3}{c}{\textbf{GAIA}} & \multicolumn{3}{c}{\textbf{WebWalkerQA}} & {\multirow{2}{*}{\textbf{Avg.}}}\\
             \cmidrule(l{0.5em}r{0.5em}){3-5} \cmidrule(l{0.5em}r{0.5em}){6-8}
             & & {\textbf{Level 1}} & {\textbf{Level 2}} & {\textbf{Level 3}} & {\textbf{Easy}} & {\textbf{Medium}} & {\textbf{Hard}} & \\
             \midrule
             Base agent &
             {\small 79.3} & {\small 68.4} & {\small 61.6} & {\small \underline{41.7}} & {\small 61.8} & {\small 59.5} & {\small 68.0} & {\small 62.9} \\
             \cmidrule{0-8}
             Confidence &
             {\small 61.3} & {\small 60.7} & {\small 47.5} & {\small 25.0} & {\small 63.4} & {\small 62.0} & {\small 64.9} & {\small 55.0} \\
              Relevance &
             {\small 81.3} & {\small \textbf{70.1}} & {\small 63.5} & {\small 33.3} & {\small \underline{66.7}} & {\small 62.8} & {\small 66.7} & {\small 63.5} \\
             Verbal-progress &
             {\small \textbf{82.3}} & {\small \underline{69.2}} & {\small 60.9} & {\small \underline{41.7}} & {\small 63.4} & {\small \underline{63.8}} & {\small 68.4} & {\small \underline{64.2}} \\
             \cmidrule{0-8}
             GenPRM-7B &
             {\small 79.0} & {\small \textbf{70.1}} & {\small \underline{64.1}} & {\small 38.9} & {\small 60.2} & {\small \underline{63.8}} & {\small \underline{68.9}} & {\small 63.6} \\
             Web-Shepherd-8B &
             {\small 79.7} & {\small \underline{69.2}} & {\small 61.5} & {\small 36.1} & {\small 62.6} & {\small 62.3} & {\small 67.1} & {\small 62.6} \\
             StepWiser &
             {\small 81.0} & {\small \textbf{70.1}} & {\small 60.9} & {\small 36.1} & {\small 65.0} & {\small 61.8} & {\small 64.9} & {\small 62.8} \\
             \cmidrule{0-8}
             \cellcolor{gg}\ours &
             \cellcolor{gg}{\small \underline{81.7}} & \cellcolor{gg}{\small \underline{69.2}} & \cellcolor{gg}{\small \textbf{65.4}} & \cellcolor{gg}{\small \textbf{44.5}} & \cellcolor{gg}{\small \textbf{70.7}} & \cellcolor{gg}{\small \textbf{65.9}} & \cellcolor{gg}{\small \textbf{70.1}} & \cellcolor{gg}{\small \textbf{66.8}} \\
             \bottomrule
        \end{tabular}
    \caption{Comparison of step quality evaluation methods on Tongyi DeepResearch-30B-A3B across information-seeking benchmarks. We adopt the LLM-as-Judge (LasJ) metric and report Avg@3. The best and the second best results are in \textbf{bold} and \underline{underline}, respectively. The results show that \ours enhances the performance of strong information-seeking agents.
    }
    \label{tab:deepresearch_eval}
\end{table*}

\section{Experiments}\label{sec:experiments}

\paragraph{Models.}
To evaluate the efficacy and generalizability of \ours, we use three distinct LLMs: Qwen3-32B~\citep{qwen3}, an open-source model with strong reasoning capability; Gemini-2.5-Flash~\citep{gemini}, a closed-source frontier model; and Tongyi DeepResearch-30B-A3B~\citep{tongyi2025deepresearch}, a recently developed agent specifically optimized for long-horizon information-seeking tasks. 
We instantiate ReAct-based agents using these LLMs.

\paragraph{Benchmarks and Metrics.}
Following recent work on long-horizon information-seeking~\citep{Gao2025beyondtenturns,Li2025webthinker,tongyi2025deepresearch}, we assess PRM-guided reasoning on three benchmarks: FRAMES~\citep{Krishna2025frames}, GAIA~\citep{Mialon2024gaia}, and WebWalkerQA~\citep{Wu2025webwalker}.
We evaluate both Qwen3-32B and DeepResearch agent across all three benchmarks, while Gemini-2.5-Falsh is evaluated on GAIA.
Further explanations of these benchmarks are provided in~\Cref{appendix:Implementation Details}.
Following past work~\citep{Gao2025beyondtenturns,Li2025webthinker,tongyi2025deepresearch}, we adopt the LLM-as-Judge (LasJ) paradigm to measure benchmark performance, which is a standard approach in long-horizon information-seeking research.
We use GPT-5 to judge the correctness of final answers. All results are reported using Avg@3, defined as the mean accuracy over three independent runs.

\paragraph{Baselines.}
We compare \ours against three categories: \textbf{(1) Base agent}: the unguided LLM, serving as a reference point for test-time improvement.
\textbf{(2) Intrinsic reasoning heuristics}: widely used reasoning quality heuristics, including confidence~\citep{Ghasemabadi2025guidedbygut}, relevance~\citep{Wan2025relevance}, and verbal-progress.
\textbf{(3) Existing PRMs}: GenPRM-7B~\citep{Zhao2025genprm}, Web-Shepherd-8B~\citep{Chae2025webshepherd}, StepWiser~\citep{Xiong2025stepwiser}. We reimplement StepWiser, which outputs binary judgment, using the same annotations and backbone model as \ours to provide a controlled comparison (c.f. ~\Cref{appendix:Implementation Details}).
This lets us isolate the contributions of dense comparative scoring and compact trajectory summarization.

\paragraph{Implementation Details.}
We use Qwen3-32B to annotate information gain score and summary for each step.
We initialize \ours with Qwen3-4B, training with one epoch of SFT for the summarization objective, followed by a period of GRPO training for the scoring objective.
This SFT-GRPO cycle is repeated for a total of X iterations.
At test-time, the LLM policy generates $n=4$ next steps which are evaluated via the baselines and \ours, and the highest scoring candidate is selected. 
Further details can be found in~\Cref{appendix:Implementation Details}.

\subsection{Results and Discussion}\label{sec:experiments:exp_results}
\paragraph{\ours substantially outperforms existing PRMs on foundation models.} 
\Cref{tab:qwen3_eval} shows \ours consistently achieves substantial gains over the Qwen3-32B base agent across all long-horizon information-seeking benchmarks.
Stepwiser yields only a 1.5\% absolute average accuracy gain compared to 9.3\% with \ours despite using the same data, highlighting the limitations of the coarse supervision binary correctness provides. 
Yet, expressiveness alone is insufficient: even baselines with richer outputs -- Verbal-progress (scalar) and Web-Shepherd (multi-item checklists) -- add only marginal gains of 0.2\% and 0.5\%, respectively. 
In contrast, \ours is trained to derive multi-factor analyses before outputting a dense comparative score grounded in both information-gain estimation and pairwise preference learning, letting it identify subtle yet key quality differences and select the most informative steps.

\begin{table}[t]
    \centering
    \begin{minipage}{0.48\textwidth}
    \centering
    \small
        \renewcommand{\arraystretch}{1.}
        \setlength{\tabcolsep}{5.5pt}
        \begin{tabular}{l c c c c}
             \toprule
             {\multirow{2}{*}{\textbf{Method}}} & \multicolumn{3}{c}{\textbf{GAIA}} & {\multirow{2}{*}{\textbf{Avg.}}}\\
             \cmidrule(l{0.5em}r{0.5em}){2-4} 
              & {\textbf{Level 1}} & {\textbf{Level 2}} & {\textbf{Level 3}} & \\
             \midrule
             Base agent &
             {\small 58.1} & {\small 42.3} & {\small \underline{19.5}} & {\small 40.0} \\
             \cmidrule{0-4}
              Relevance &
             {\small 58.1} & {\small 44.9} & {\small \underline{19.5}} & {\small 40.8} \\
             Verbal-progress &
             {\small \underline{60.7}} & {\small 44.2} & {\small 16.7} & {\small 40.5} \\
             \cmidrule{0-4}
             GenPRM-7B &
             {\small 56.4} & {\small 44.9} & {\small 11.1} & {\small 37.5} \\
             Web-Shepherd-8B &
             {\small 59.8} & {\small \textbf{46.2}} & {\small 16.7} & {\small 40.9} \\
             StepWiser &
             {\small \underline{60.7}} & {\small 44.3} & {\small \underline{19.5}} & {\small \underline{41.5}} \\
             \cmidrule{0-4}
             \cellcolor{gg} \ours &
             \cellcolor{gg}{\small \textbf{61.5}} & \cellcolor{gg}{\small \underline{45.5}} & \cellcolor{gg}{\small \textbf{25.0}} & \cellcolor{gg}{\small \textbf{44.0}} \\
             \bottomrule
        \end{tabular}
    \caption{\textbf{\ours shows strong generalization to the frontier LLM (Gemini).} We adopt the LLM-as-Judge (LasJ) metric and report Avg@3 comparing with other step quality evaluation methods on Gemini-2.5-Flash.
    }
    \label{tab:gemini_eval}
    \end{minipage}
\end{table}

\paragraph{\ours improves highly performant informa-tion-seeking agents.}
In~\Cref{tab:deepresearch_eval}, we evaluate \ours on DeepResearch-30B-A3B, a specialized agent optimized for long-horizon information-seeking tasks.
The results show that adding \ours to this strong information-seeking agent consistently achieves performance gains across benchmarks, surpassing the base agent by 3.9\% absolute average accuracy, while no other baselines come close to achieving notable improvements. 
Notably, \ours improves on the challenging subsets, such as GAIA Level 3 and WebWalkerQA Hard. 
Moreover, on GAIA, \ours lifts DeepResearch-30B-A3B from 61.9\% to 64.4\% average accuracy, enabling the 30B agent augmented with the 4B PRM to reach competitive performance with OpenAI DeepResearch (67.4\%) and surpass DeepSeek-V3.1-671B (63.1\%) \citep{tongyi2025deepresearch}.\footnote{See~\Cref{appendix:Implementation Details} for frontier model results reporting details.}
This test-time scaling approach is substantially more efficient than scaling the base model; it achieves frontier-level performance while requiring significantly less memory and computational overhead than deploying the 20-times-larger 671B model.
Crucially, while specialized information-seeking agents~\citep{Wu2025webdancer,Li2025websailor,Li2025webthinker} require resource-intensive fine-tuning with massive datasets (10k-100k+ samples) and costly online RL (tool interactions and multi-step rollouts)~\citep{Gao2025beyondtenturns}, \ours achieves these gains using a 4B model that either does not require a large dataset (2k+ pair samples) or tool interactions and long-horizon rollouts during training.
This demonstrates that step-level guidance provided by \ours is a cost-effective strategy to push the performance limits of even highly optimized information-seeking agents.

\paragraph{\ours also generalizes to frontier LLMs.}
To further demonstrate the versatility of our approach, we use the closed-source Gemini-2.5-Flash as the LLM agent, as shown in~\Cref{tab:gemini_eval}.
\ours provides 4.0\% absolute average accuracy gain, whereas the second-best method improves performance by only 1.5\%. On the most challenging subset, GAIA Level 3, \ours yields the largest improvement among all baselines (+5.5\%), showing its strength on long-horizon reasoning tasks.
Overall, our results indicate that \ours provides effective test-time guidance, improving the information-seeking behavior of both open-source LLMs, closed-source LLMs, and information-seeking agents, which shows strong versatility and generalizability without modifying or retraining underlying LLMs.

\subsection{Analysis and Ablations \label{sec:subsec:analysis}}

\paragraph{Summarization ability contributes to better scoring ability.} 
To validate the effectiveness of compressed representations for accurate step-level scoring, we compare our summarization approach against several alternatives: providing the most recent one, two, or four steps as input context ($H_{-1:}$, $H_{-2:}$, $H_{-4:}$), and the full trajectory ($H_{t}$).
We evaluate on Qwen3-32B across FRAMES and GAIA. Since performance on GAIA Level 3 is low (see~\Cref {tab:qwen3_eval}), we use Levels 1 and 2.
Results in~\Cref{tab:history_context_ablation} show that our summarization approach achieves the best or second-best performance across benchmarks, outperforming the full raw-history baseline by 7.7\% absolute average accuracy.

Notably, extending raw history does not improve performance. $H_{-2:}$ outperforms both $H_{-1:}$ (insufficient context), $H_{-4:}$, and $H_{t}$ (excessive, noisy context).
This confirms that long histories introduce noise and irrelevant information that distract step-level evaluation.
In contrast, our summarization compresses entire trajectories into compact representations that preserve key information while filtering noise, enabling accurate scoring even as trajectories grow arbitrarily long. 

\begin{table}[t]
    \centering
    \begin{minipage}{0.48\textwidth}
    \centering
    \small
        \renewcommand{\arraystretch}{1.0}
        \setlength{\tabcolsep}{5.5pt}
        \begin{tabular}{l c c c c}
             \toprule
             {\multirow{2}{*}{\textbf{Input Context}}} & {\multirow{2}{*}{\textbf{FRAMES}}} & \multicolumn{2}{c}{\textbf{GAIA}} & {\multirow{2}{*}{\textbf{Avg.}}}\\
             \cmidrule(l{0.5em}r{0.5em}){3-4} 
              & & {\textbf{Level 1}} & {\textbf{Level 2}} & \\
             \midrule
             $H_{-1:}$ &
             {\small 56.3} & {\small 44.5} & {\small 25.7} & {\small 42.2} \\
             $H_{-2:}$ &
             {\small \textbf{61.0}} & {\small 44.5} & {\small 26.9} & {\small 44.1} \\
             $H_{-4:}$ &
             {\small 57.0} & {\small 37.6} & {\small 25.0} & {\small 39.9} \\
             $H_t$ &
             {\small 55.7} & {\small 38.5} & {\small 24.4} & {\small 39.5} \\
              \cmidrule{0-4}
             $h_t$ (Ours) &
             {\small 58.7} & {\small \textbf{49.6}} & {\small \textbf{33.3}} & {\small \textbf{47.2}} \\
             \bottomrule
        \end{tabular}
    \caption{\textbf{Effectiveness of context compression.} Comparison of input context representations for PRM on Qwen3-32B across information-seeking tasks. $H_{-1:}, H_{-2:}$, and $H_{-4:}$ provide the most recent one, two, and four trajectory steps from the full trajectory $H_t$, while $h_t$ uses the trajectory summary from \ours. Our approach ($h_t$) shows better scoring ability by retaining essential information for step evaluation.}
    \label{tab:history_context_ablation}
    \end{minipage}
\end{table}

\paragraph{Complementary rewards improve step-level evaluation.} 
We analyze the contribution of each reward component in~\Cref{eq:final_reward}. 
Following the setup from the previous ablation, we evaluate on Qwen3-32B across FRAMES, GAIA Levels 1, and 2.
As shown in~\Cref{tab:reward_ablation}, combining the score and comparison rewards ($r_s+r_c$) yields substantially better performance than using either component alone, leading to 2.0\% and 3.1\% absolute average accuracy gains compared to using the score reward ($r_s$) and comparison reward ($r_c$), respectively.
This indicates that information-gain estimation and preference prediction capture complementary aspects of the quality of a trajectory step, underscoring the benefit of our pairwise annotation strategy over prior work that labels individual steps in isolation~\citep{Xiong2025stepwiser,Wang2024mathshepherd}. 
Furthermore, incorporating the adaptive weight ($r_s+w\cdot r_c$) yields 1.0\% additional absolute average accuracy gain over the naive combination.
This is because the adaptive weight mitigates noise in preference pairs.
Pairs with small information gain score differences are inherently noisier, as they may reflect annotation variance rather than true quality differences.
Thus, these pairs receive lower weights, while pairs with clear margins are weighted higher, leading to more stable learning.
Overall, adaptive weighting provides a simple and cost-efficient way of leveraging existing annotated preference pairs.

\begin{table}[t]
    \centering
    \begin{minipage}{0.48\textwidth}
    \centering
    \small
        \renewcommand{\arraystretch}{1.0}
        \setlength{\tabcolsep}{1.0pt}
        \begin{tabular}{l c c c c}
             \toprule
             {\multirow{2}{*}{\textbf{Reward Design}}} & {\multirow{2}{*}{\textbf{FRAMES}}} & \multicolumn{2}{c}{\textbf{GAIA}} & {\multirow{2}{*}{\textbf{Avg.}}}\\
            \cmidrule(l{0.5em}r{0.5em}){3-4} 

              & & {\textbf{Level 1}} & {\textbf{Level 2}} & \\
             \midrule
             $r=r_s$ (score-only) &
             {\small 57.0} & {\small 43.6} & {\small 32.0} & {\small 44.2} \\
             $r=r_c$ (comparison-only) &
             {\small 58.7} & {\small 41.9} & {\small 28.8} & {\small 43.1} \\
             $r=r_s+r_c$ (combination) &
             {\small \textbf{60.3}} & {\small 47.0} & {\small 31.4} & {\small 46.2} \\
             \cmidrule{0-4}
             $r=r_s+w\cdot r_c$ (Ours) &
             {\small 58.7} & {\small \textbf{49.6}} & {\small \textbf{33.3}} & {\small \textbf{47.2}} \\
             \bottomrule
        \end{tabular}
    \caption{\textbf{Impact of reward components.} Experiments with reward components on PRM performance across information-seeking tasks evaluated with Qwen3-32B. Combining the score reward with comparison reward ($r_s+r_c$) leads to better step evaluation, with further improvement when mitigating noise in preference pairs through adaptive weight ($r_s+w\cdot r_c$).
    }
    \label{tab:reward_ablation}
    \end{minipage}
\end{table}

\begin{figure}[t]
    \centering
    \includegraphics[width=\linewidth]{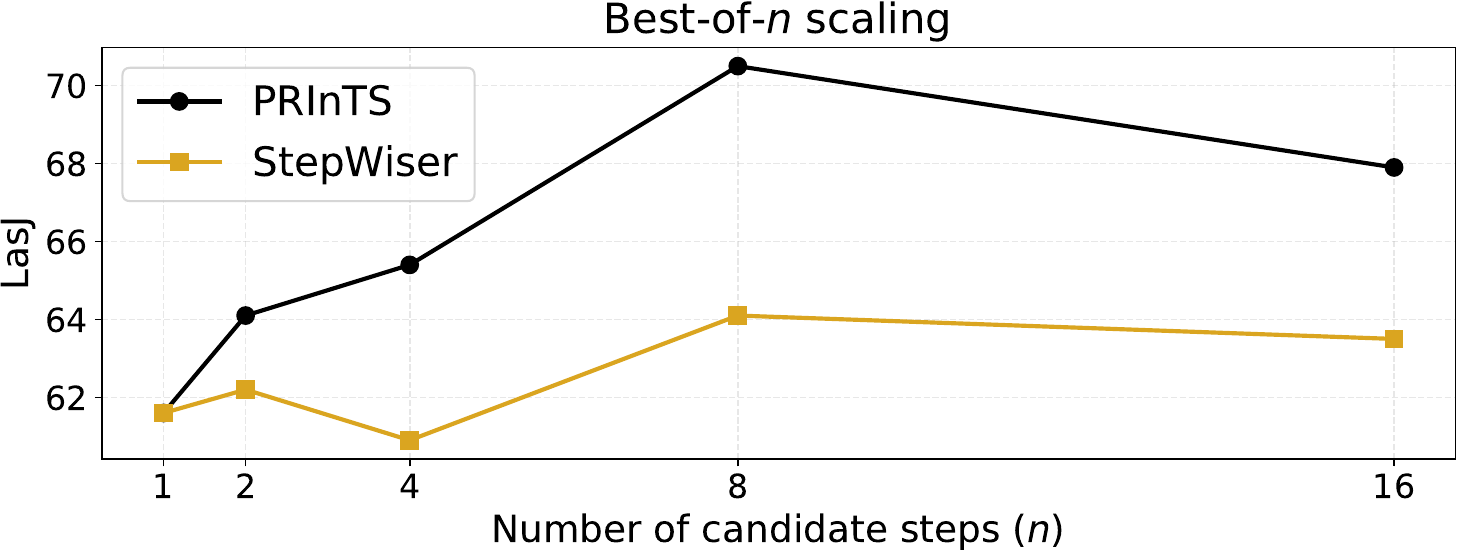}
    \par
    \caption{\textbf{Scaling test-time compute.} Best-of-$n$ test-time scaling results on GAIA Level 2 using Tongyi DeepResearch-30B-A3B. \ours benefits from additional test-time compute by identifying higher-quality steps from $n$ candidates.
    }
    \label{fig:n_scale}
\end{figure}

\paragraph{\ours scales effectively with test-time compute.}
In order to evaluate how \ours benefits from additional test-time compute, we conduct best-of-$n$ scaling experiments with varying numbers of candidate steps ($n\in\{1,2,4,8,16\}$) on GAIA Level 2 using Qwen3-32B, as shown in~\Cref{fig:n_scale}.
\ours exhibits strong scaling behavior, achieving 2.5\%, 3.8\%, 8.9\% absolute accuracy gains at $n\!=\!2,\;4,\;8$, respectively, demonstrating that \ours reliably identifies higher-quality steps within large candidate sets.
However, performance declines at $n=16$. We attribute this to the expanded candidate pool. A larger $n$ increases the likelihood of generating steps that superficially appear to resolve uncertainty or offer high information gain.
Hence, it becomes increasingly prone to selecting these seemingly informative exploratory steps over candidates that correctly output the final answer.
Consequently, the agent abandons shorter, successful trajectories and continues exploring until it reaches the maximum step budget, failing to output an answer despite having generated one earlier.
In contrast, StepWiser provides only marginal and inconsistent improvements under scaling.
This difference in scaling efficiency further validates that \ours's design of information-gain estimation and preference prediction captures subtle quality differences between steps, enabling fine-grained guidance for long-horizon information-seeking.

\section{Conclusion}
In this paper, we introduce \ours, a generative PRM for long-horizon information seeking.
\ours unifies information gain scoring with recursive trajectory summarization, enabling fine-grained step-level evaluation under rapidly accumulating agent context.
To equip \ours with these dual abilities, we construct preference step pairs with information gain scores and summaries and jointly train \ours by alternating supervised fine-tuning for summarization and reinforcement learning for scoring.
We test \ours on three strong and distinct agents and demonstrate that \ours consistently enhances their information-seeking abilities, showcasing its high versatility. 
\ours also improves frontier information-seeking agents, showing that test-time guidance is a powerful complement to agent fine-tuning and robust to changes in underlying models.

\section*{Limitations}
While our pipeline automates the annotation of information gain scores and summaries, the underlying QA samples we collected are focused on English queries. Expanding coverage to multilingual contexts remains a promising direction to further enhance \ours's versatility.
Another compelling avenue for future research is to employ \ours as a dense reward signal for training information-seeking agents via reinforcement learning. We leave this exploration for future work due to its significant computational demands, focusing instead on demonstrating test-time improvements.

\section*{Ethical Considerations}
In this work, we suggest \ours that can improve the information-seeking ability of agent systems by providing finer-grained evaluation of intermediate reasoning and tool interactions, potentially reducing erroneous or inefficient information-seeking behavior.
By improving agents' ability to gather and synthesize information accurately, our work could benefit applications such as scientific research and knowledge discovery.
We do not foresee any ethical implications beyond those applicable to agentic research systems more generally.

\section*{Acknowledgments}
This work was supported by NSF-AI Engage Institute DRL-2112635, NSF-CAREER Award 1846185, DARPA ECOLE Program No. HR00112390060, Capital One Research Award, Apple PhD Fellowship, NDSEG PhD Fellowship. The views contained in this article are those of the authors and not of the funding agency.

\bibliography{custom}

@string{iclr = "Proceedings of the International Conference on Learning Representations (ICLR)"}

@string{neurips = "Advances in Neural Information Processing Systems (NeurIPS)"}

@string{emnlp = "Proceedings of the Conference on Empirical Methods in Natural Language Processing (EMNLP)"}

@string{acl = "Proceedings of the Association for Computational Linguistics (ACL)"}

@string{naacl = "Proceedings of the North American Chapter of the Association for Computational Linguistics (NAACL)"}

@string{emnlp_findings = "Findings of the Conference on Empirical Methods in Natural Language Processing (EMNLP)"}

@article{Yang2025swesmith,
  author       = {John Yang and
                  Kilian Leret and
                  Carlos E. Jimenez and
                  Alexander Wettig and
                  Kabir Khandpur and
                  Yanzhe Zhang and
                  Binyuan Hui and
                  Ofir Press and
                  Ludwig Schmidt and
                  Diyi Yang},
  title        = {SWE-smith: Scaling Data for Software Engineering Agents},
  journal      = {arXiv preprint arXiv:2504.21798},
  year         = {2025},
  url          = {https://doi.org/10.48550/arXiv.2504.21798},
}

@article{Pan2024swegym,
  author       = {Jiayi Pan and
                  Xingyao Wang and
                  Graham Neubig and
                  Navdeep Jaitly and
                  Heng Ji and
                  Alane Suhr and
                  Yizhe Zhang},
  title        = {Training Software Engineering Agents and Verifiers with SWE-Gym},
  journal      = {arXiv preprint arXiv:2412.21139},
  year         = {2024},
  url          = {https://doi.org/10.48550/arXiv.2412.21139},
}

@article{Liu20025agenticmath,
  author       = {Xianyang Liu and
                  Yilin Liu and
                  Shuai Wang and
                  Hao Cheng and
                  Andrew Estornell and
                  Yuzhi Zhao and
                  Jiaheng Wei},
  title        = {AgenticMath: Enhancing {LLM} Reasoning via Agentic-based Math Data
                  Generation},
  journal      = {arXiv preprint arXiv:2510.19361},
  year         = {2025},
  url          = {https://doi.org/10.48550/arXiv.2510.19361},
}

@article{Liu2025mmagent,
  author       = {Fan Liu and
                  Zherui Yang and
                  Cancheng Liu and
                  Tianrui Song and
                  Xiaofeng Gao and
                  Hao Liu},
  title        = {MM-Agent: {LLM} as Agents for Real-world Mathematical Modeling Problem},
  journal      = {arXiv preprint arXiv:2505.14148},
  year         = {2025},
  url          = {https://doi.org/10.48550/arXiv.2505.14148},
}

@inproceedings{Prasad2023receval,
  author       = {Archiki Prasad and
                  Swarnadeep Saha and
                  Xiang Zhou and
                  Mohit Bansal},
  title        = {ReCEval: Evaluating Reasoning Chains via Correctness and Informativeness},
  booktitle    = emnlp,
  year         = {2023},
}

@inproceedings{Rao2018askgoodquestion,
  author       = {Sudha Rao and
                  Hal Daum{\'{e}} III},
  title        = {Learning to Ask Good Questions: Ranking Clarification Questions using
                  Neural Expected Value of Perfect Information},
  booktitle    = acl,
  year         = {2018},
}

@article{Gao2025beyondtenturns,
  author       = {Jiaxuan Gao and
                  Wei Fu and
                  Minyang Xie and
                  Shusheng Xu and
                  Chuyi He and
                  Zhiyu Mei and
                  Banghua Zhu and
                  Yi Wu},
  title        = {Beyond Ten Turns: Unlocking Long-Horizon Agentic Search with Large-Scale
                  Asynchronous {RL}},
  journal      = {arXiv preprint arXiv:2508.07976},
  year         = {2025},
  url          = {https://doi.org/10.48550/arXiv.2508.07976},
}

@article{Li2025webthinker,
  author       = {Xiaoxi Li and
                  Jiajie Jin and
                  Guanting Dong and
                  Hongjin Qian and
                  Yutao Zhu and
                  Yongkang Wu and
                  Ji{-}Rong Wen and
                  Zhicheng Dou},
  title        = {WebThinker: Empowering Large Reasoning Models with Deep Research Capability},
  journal      = {arXiv preprint arXiv:2504.21776},
  year         = {2025},
  url          = {https://doi.org/10.48550/arXiv.2504.21776},
}

@article{Wang2025eat,
      title={Entropy After $\langle \texttt{/Think} \rangle$ for reasoning model early exiting}, 
      author={Xi Wang and James McInerney and Lequn Wang and Nathan Kallus},
      year={2025},
      journal      = {arXiv preprint arXiv:2509.26522},
      url={https://arxiv.org/abs/2509.26522}, 
}

@inproceedings{Yao2023react,
  author       = {Shunyu Yao and
                  Jeffrey Zhao and
                  Dian Yu and
                  Nan Du and
                  Izhak Shafran and
                  Karthik R. Narasimhan and
                  Yuan Cao},
  title        = {ReAct: Synergizing Reasoning and Acting in Language Models},
  booktitle    = iclr,
  year         = {2023},
}

@inproceedings{Wu2025mindmap,
  author       = {Junde Wu and
                  Jiayuan Zhu and
                  Yuyuan Liu and
                  Min Xu and
                  Yueming Jin},
  title        = {Agentic Reasoning: {A} Streamlined Framework for Enhancing {LLM} Reasoning
                  with Agentic Tools},
  booktitle    = acl,
  year         = {2025},
}

@article{kang2025acon,
  title={ACON: Optimizing Context Compression for Long-horizon LLM Agents}, 
  author={Minki Kang and Wei-Ning Chen and Dongge Han and Huseyin A. Inan and Lukas Wutschitz and Yanzhi Chen and Robert Sim and Saravan Rajmohan},
  journal      = {arXiv preprint arXiv:2510.00615},
  year         = {2025},
  url          = {https://arxiv.org/abs/2510.00615},
}

@article{Ye2025agentfold,
      title={AgentFold: Long-Horizon Web Agents with Proactive Context Management}, 
      author={Rui Ye and Zhongwang Zhang and Kuan Li and Huifeng Yin and Zhengwei Tao and Yida Zhao and Liangcai Su and Liwen Zhang and Zile Qiao and Xinyu Wang and Pengjun Xie and Fei Huang and Siheng Chen and Jingren Zhou and Yong Jiang},
      year={2025},
      journal      = {arXiv preprint arXiv:2510.24699},
      url={https://arxiv.org/abs/2510.24699}, 
}

@article{yen2025lostmaze,
      title={Lost in the Maze: Overcoming Context Limitations in Long-Horizon Agentic Search}, 
      author={Howard Yen and Ashwin Paranjape and Mengzhou Xia and Thejas Venkatesh and Jack Hessel and Danqi Chen and Yuhao Zhang},
      year={2025},
      journal      = {arXiv preprint arXiv:2510.18939},
      url={https://arxiv.org/abs/2510.18939}, 
}

@article{tang2025longrm,
      title={LongRM: Revealing and Unlocking the Context Boundary of Reward Modeling}, 
      author={Zecheng Tang and Baibei Ji and Quantong Qiu and Haitian Wang and Xiaobo Liang and Juntao Li and Min Zhang},
      year={2025},
      journal      = {arXiv preprint arXiv:2510.06915},
      url={https://arxiv.org/abs/2510.06915}, 
}

@article{Wang2024selftaughtevaluators,
  author       = {Tianlu Wang and
                  Ilia Kulikov and
                  Olga Golovneva and
                  Ping Yu and
                  Weizhe Yuan and
                  Jane Dwivedi{-}Yu and
                  Richard Yuanzhe Pang and
                  Maryam Fazel{-}Zarandi and
                  Jason Weston and
                  Xian Li},
  title        = {Self-Taught Evaluators},
  journal      = {arXiv preprint arXiv:2408.02666},
  year         = {2024},
  url          = {https://doi.org/10.48550/arXiv.2408.02666},
}

@article{Whitehouse2025j1,
  author       = {Chenxi Whitehouse and
                  Tianlu Wang and
                  Ping Yu and
                  Xian Li and
                  Jason Weston and
                  Ilia Kulikov and
                  Swarnadeep Saha},
  title        = {{J1:} Incentivizing Thinking in LLM-as-a-Judge via Reinforcement Learning},
  journal      = {arXiv preprint arXiv:2505.10320},
  year         = {2025},
  url          = {https://doi.org/10.48550/arXiv.2505.10320},
}

@inproceedings{Kim2024prospector,
  author       = {Byoungjip Kim and
                  Youngsoo Jang and
                  Lajanugen Logeswaran and
                  Geon{-}Hyeong Kim and
                  Yu Jin Kim and
                  Honglak Lee and
                  Moontae Lee},
  editor       = {Yaser Al{-}Onaizan and
                  Mohit Bansal and
                  Yun{-}Nung Chen},
  title        = {Prospector: Improving {LLM} Agents with Self-Asking and Trajectory Ranking},
  booktitle    = emnlp_findings,
  year         = {2024},
}

@article{Zou2025reasonflux,
  author       = {Jiaru Zou and
                  Ling Yang and
                  Jingwen Gu and
                  Jiahao Qiu and
                  Ke Shen and
                  Jingrui He and
                  Mengdi Wang},
  title        = {ReasonFlux-PRM: Trajectory-Aware PRMs for Long Chain-of-Thought Reasoning
                  in LLMs},
  journal      = {arXiv preprint arXiv:2506.18896},
  year         = {2025},
  url          = {https://doi.org/10.48550/arXiv.2506.18896},
}

@article{Xiong2025stepwiser,
  author       = {Wei Xiong and
                  Wenting Zhao and
                  Weizhe Yuan and
                  Olga Golovneva and
                  Tong Zhang and
                  Jason Weston and
                  Sainbayar Sukhbaatar},
  title        = {StepWiser: Stepwise Generative Judges for Wiser Reasoning},
  journal      = {arXiv preprint arXiv:2508.19229},
  year         = {2025},
  url          = {https://doi.org/10.48550/arXiv.2508.19229},
}

@article{Ton2024infogainprm,
  author       = {Jean{-}Francois Ton and
                  Muhammad Faaiz Taufiq and
                  Yang Liu},
  title        = {Understanding Chain-of-Thought in LLMs through Information Theory},
  journal      = {arXiv preprint arXiv:2411.11984},
  year         = {2024},
  url          = {https://doi.org/10.48550/arXiv.2411.11984},
}

@article{Chae2025webshepherd,
  author       = {Hyungjoo Chae and
                  Sunghwan Kim and
                  Junhee Cho and
                  Seungone Kim and
                  Seungjun Moon and
                  Gyeom Hwangbo and
                  Dongha Lim and
                  Minjin Kim and
                  Yeonjun Hwang and
                  Minju Gwak and
                  Dongwook Choi and
                  Minseok Kang and
                  Gwanhoon Im and
                  ByeongUng Cho and
                  Hyojun Kim and
                  Jun Hee Han and
                  Taeyoon Kwon and
                  Minju Kim and
                  Beong{-}woo Kwak and
                  Dongjin Kang and
                  Jinyoung Yeo},
  title        = {Web-Shepherd: Advancing PRMs for Reinforcing Web Agents},
  journal      = {arXiv preprint arXiv:2505.15277},
  year         = {2025},
  url          = {https://doi.org/10.48550/arXiv.2505.15277},
}

@inproceedings{Wang2024mathshepherd,
  author       = {Peiyi Wang and
                  Lei Li and
                  Zhihong Shao and
                  Runxin Xu and
                  Damai Dai and
                  Yifei Li and
                  Deli Chen and
                  Yu Wu and
                  Zhifang Sui},
  title        = {Math-Shepherd: Verify and Reinforce LLMs Step-by-step without Human Annotations},
  booktitle    = acl,
  year         = {2024},
}

@article{Zhao2025genprm,
  author       = {Jian Zhao and
                  Runze Liu and
                  Kaiyan Zhang and
                  Zhimu Zhou and
                  Junqi Gao and
                  Dong Li and
                  Jiafei Lyu and
                  Zhouyi Qian and
                  Biqing Qi and
                  Xiu Li and
                  Bowen Zhou},
  title        = {GenPRM: Scaling Test-Time Compute of Process Reward Models via Generative
                  Reasoning},
  journal      = {arXiv preprint arXiv:2504.00891},
  year         = {2025},
  url          = {https://doi.org/10.48550/arXiv.2504.00891},
}

@article{Deng2025atomsearcher,
  author       = {Yong Deng and
                  Guoqing Wang and
                  Zhenzhe Ying and
                  Xiaofeng Wu and
                  Jinzhen Lin and
                  Wenwen Xiong and
                  Yuqin Dai and
                  Shuo Yang and
                  Zhanwei Zhang and
                  Qiwen Wang and
                  Yang Qin and
                  Yuan Wang and
                  Quanxing Zha and
                  Sunhao Dai and
                  Changhua Meng},
  title        = {Atom-Searcher: Enhancing Agentic Deep Research via Fine-Grained Atomic
                  Thought Reward},
  journal      = {arXiv preprint arXiv:2508.12800},
  year         = {2025},
  url          = {https://doi.org/10.48550/arXiv.2508.12800},
}

@article{He2025generativeprm,
  author       = {Tao He and
                  Rongchuan Mu and
                  Lizi Liao and
                  Yixin Cao and
                  Ming Liu and
                  Bing Qin},
  title        = {Good Learners Think Their Thinking: Generative {PRM} Makes Large Reasoning Model More Efficient Math Learner},
  journal      = {arXiv preprint arXiv:2507.23317},
  volume       = {abs/2507.23317},
  year         = {2025},
  url          = {https://doi.org/10.48550/arXiv.2507.23317},
}

@article{Zhou2025finprm,
  author       = {Yuanchen Zhou and
                  Shuo Jiang and
                  Jie Zhu and
                  Junhui Li and
                  Lifan Guo and
                  Feng Chen and
                  Chi Zhang},
  title        = {Fin-PRM: {A} Domain-Specialized Process Reward Model for Financial Reasoning in Large Language Models},
  journal      = {arXiv preprint arXiv:2508.15202},
  year         = {2025},
  url          = {https://doi.org/10.48550/arXiv.2508.15202},
}

@article{Choudhury2025agentprm,
  author       = {Sanjiban Choudhury},
  title        = {Process Reward Models for {LLM} Agents: Practical Framework and Directions},
  journal      = {arXiv preprint arXiv:2502.10325},
  year         = {2025},
  url          = {https://doi.org/10.48550/arXiv.2502.10325},
}

@inproceedings{Setlur2025pav,
  author       = {Amrith Setlur and
                  Chirag Nagpal and
                  Adam Fisch and
                  Xinyang Geng and
                  Jacob Eisenstein and
                  Rishabh Agarwal and
                  Alekh Agarwal and
                  Jonathan Berant and
                  Aviral Kumar},
  title        = {Rewarding Progress: Scaling Automated Process Verifiers for {LLM} Reasoning},
  booktitle    = iclr,
  year         = {2025},
}

@article{Ghasemabadi2025guidedbygut,
  author       = {Amirhosein Ghasemabadi and
                  Keith G. Mills and
                  Baochun Li and
                  Di Niu},
  title        = {Guided by Gut: Efficient Test-Time Scaling with Reinforced Intrinsic Confidence},
  journal      = {arXiv preprint arXiv:2505.20325},
  year         = {2025},
  url          = {https://doi.org/10.48550/arXiv.2505.20325},
}

@article{Prabhudesai2025maximizeconfidence,
  author       = {Mihir Prabhudesai and
                  Lili Chen and
                  Alex Ippoliti and
                  Katerina Fragkiadaki and
                  Hao Liu and
                  Deepak Pathak},
  title        = {Maximizing Confidence Alone Improves Reasoning},
  journal      = {arXiv preprint arXiv:2505.22660},
  year         = {2025},
  url          = {https://doi.org/10.48550/arXiv.2505.22660},
}

@article{Fu2025deepthinkwconfidence,
  author       = {Yichao Fu and
                  Xuewei Wang and
                  Yuandong Tian and
                  Jiawei Zhao},
  title        = {Deep Think with Confidence},
  journal      = {arXiv preprint arXiv:2508.15260},
  year         = {2025},
  url          = {https://doi.org/10.48550/arXiv.2508.15260},
}

@inproceedings{Wan2025relevance,
  author       = {Guangya Wan and
                  Yuqi Wu and
                  Jie Chen and
                  Sheng Li},
  title        = {Reasoning Aware Self-Consistency: Leveraging Reasoning Paths for Efficient
                  {LLM} Sampling},
  booktitle    = naacl,
  year         = {2025},
}

@article{Wu2025webdancer,
  author       = {Jialong Wu and
                  Baixuan Li and
                  Runnan Fang and
                  Wenbiao Yin and
                  Liwen Zhang and
                  Zhengwei Tao and
                  Dingchu Zhang and
                  Zekun Xi and
                  Yong Jiang and
                  Pengjun Xie and
                  Fei Huang and
                  Jingren Zhou},
  title        = {WebDancer: Towards Autonomous Information Seeking Agency},
  journal      = {arXiv preprint arXiv:2505.22648},
  year         = {2025},
  url          = {https://doi.org/10.48550/arXiv.2505.22648},
}

@article{Li2025websailor,
  author       = {Kuan Li and
                  Zhongwang Zhang and
                  Huifeng Yin and
                  Liwen Zhang and
                  Litu Ou and
                  Jialong Wu and
                  Wenbiao Yin and
                  Baixuan Li and
                  Zhengwei Tao and
                  Xinyu Wang and
                  Weizhou Shen and
                  Junkai Zhang and
                  Dingchu Zhang and
                  Xixi Wu and
                  Yong Jiang and
                  Ming Yan and
                  Pengjun Xie and
                  Fei Huang and
                  Jingren Zhou},
  title        = {WebSailor: Navigating Super-human Reasoning for Web Agent},
  journal      = {arXiv preprint arXiv:2507.02592},
  year         = {2025},
}

@article{Tao2025webshaper,
  author       = {Zhengwei Tao and
                  Jialong Wu and
                  Wenbiao Yin and
                  Junkai Zhang and
                  Baixuan Li and
                  Haiyang Shen and
                  Kuan Li and
                  Liwen Zhang and
                  Xinyu Wang and
                  Yong Jiang and
                  Pengjun Xie and
                  Fei Huang and
                  Jingren Zhou},
  title        = {WebShaper: Agentically Data Synthesizing via Information-Seeking Formalization},
  journal      = {arXiv preprint arXiv:2507.15061},
  year         = {2025},
  url          = {https://doi.org/10.48550/arXiv.2507.15061},
}

@misc{2025mirothinker,
    title={MiroThinker: An open-source agentic model series trained for deep research and complex, long-horizon problem solving},
    author={{MiroMind AI Team}},
    howpublished={\url{https://github.com/MiroMindAI/MiroThinker}},
    year={2025}
}

@misc{miromind2024opendata,
  title={MiroVerse V0.1: A Reproducible, Full-Trajectory, Ever-Growing Deep Research Dataset},
  author={{MiroMind Data Team}},
  year={2025},
  url={https://huggingface.co/datasets/miromind-ai/MiroVerse-v0.1}
}

@article{tongyi2025deepresearch,
      title={Tongyi DeepResearch Technical Report}, 
      author={Baixuan Li and Bo Zhang and Dingchu Zhang and Fei Huang and Guangyu Li and Guoxin Chen and Huifeng Yin and Jialong Wu and Jingren Zhou and Kuan Li and Yong Jiang et al.},
      year={2025},
      journal      = {arXiv preprint arXiv:2510.24701},
      url={https://arxiv.org/abs/2510.24701}, 
}

@article{qwen3,
  author       = {An Yang and
                  Anfeng Li and
                  Baosong Yang and
                  Beichen Zhang and
                  Binyuan Hui and
                  Bo Zheng and
                  Bowen Yu and
                  Chang Gao and
                  Chengen Huang and
                  Chenxu Lv and et al.},
  title        = {Qwen3 Technical Report},
  journal      = {arXiv preprint arXiv:2505.09388},
  year         = {2025},
  url          = {https://doi.org/10.48550/arXiv.2505.09388},
}

@article{gemini,
  author       = {{Gemini Team}},
  title        = {Gemini 2.5: Pushing the Frontier with Advanced Reasoning, Multimodality,
                  Long Context, and Next Generation Agentic Capabilities},
  journal      = {arXiv preprint arXiv:2507.06261},
  year         = {2025},
  url          = {https://doi.org/10.48550/arXiv.2507.06261},
}

@inproceedings{Mialon2024gaia,
  author       = {Gr{\'{e}}goire Mialon and
                  Cl{\'{e}}mentine Fourrier and
                  Thomas Wolf and
                  Yann LeCun and
                  Thomas Scialom},
  title        = {{GAIA:} a benchmark for General {AI} Assistants},
  booktitle    = iclr,
  year         = {2024},
}

@inproceedings{Wu2025webwalker,
  author       = {Jialong Wu and
                  Wenbiao Yin and
                  Yong Jiang and
                  Zhenglin Wang and
                  Zekun Xi and
                  Runnan Fang and
                  Linhai Zhang and
                  Yulan He and
                  Deyu Zhou and
                  Pengjun Xie and
                  Fei Huang},
  title        = {WebWalker: Benchmarking LLMs in Web Traversal},
  booktitle    = acl,
  year         = {2025},
}

@inproceedings{Krishna2025frames,
  author       = {Satyapriya Krishna and
                  Kalpesh Krishna and
                  Anhad Mohananey and
                  Steven Schwarcz and
                  Adam Stambler and
                  Shyam Upadhyay and
                  Manaal Faruqui},
  title        = {Fact, Fetch, and Reason: {A} Unified Evaluation of Retrieval-Augmented
                  Generation},
  booktitle    = acl,
  year         = {2025},
}

@article{Wei2025browsecomp,
  author       = {Jason Wei and
                  Zhiqing Sun and
                  Spencer Papay and
                  Scott McKinney and
                  Jeffrey Han and
                  Isa Fulford and
                  Hyung Won Chung and
                  Alex Tachard Passos and
                  William Fedus and
                  Amelia Glaese},
  title        = {BrowseComp: {A} Simple Yet Challenging Benchmark for Browsing Agents},
  journal      = {arXiv preprint arXiv:2504.12516},
  year         = {2025},
  url          = {https://doi.org/10.48550/arXiv.2504.12516},
}

@misc{inspect_eval,
  author = {AI Security Institute, UK},
  year = {2024},
  title = {Inspect {AI:} {Framework} for {Large} {Language} {Model}
    {Evaluations}},
  url = {https://github.com/UKGovernmentBEIS/inspect_ai},
}

@inproceedings{Su2025bright,
  author       = {Hongjin Su and
                  Howard Yen and
                  Mengzhou Xia and
                  Weijia Shi and
                  Niklas Muennighoff and
                  Han{-}yu Wang and
                  Haisu Liu and
                  Quan Shi and
                  Zachary S. Siegel and
                  Michael Tang and
                  Ruoxi Sun and
                  Jinsung Yoon and
                  Sercan {\"{O}}. Arik and
                  Danqi Chen and
                  Tao Yu},
  title        = {{BRIGHT:} {A} Realistic and Challenging Benchmark for Reasoning-Intensive Retrieval},
  booktitle    = iclr,
  year         = {2025},
}

@article{Shao2025reasonir,
  author       = {Rulin Shao and
                  Rui Qiao and
                  Varsha Kishore and
                  Niklas Muennighoff and
                  Xi Victoria Lin and
                  Daniela Rus and
                  Bryan Kian Hsiang Low and
                  Sewon Min and
                  Wen{-}tau Yih and
                  Pang Wei Koh and
                  Luke Zettlemoyer},
  title        = {ReasonIR: Training Retrievers for Reasoning Tasks},
  journal      = {arXiv preprint arXiv:2504.20595},
  year         = {2025},
  url          = {https://doi.org/10.48550/arXiv.2504.20595},
}

@article{Shao2024grpo,
  author       = {Zhihong Shao and
                  Peiyi Wang and
                  Qihao Zhu and
                  Runxin Xu and
                  Junxiao Song and
                  Mingchuan Zhang and
                  Y. K. Li and
                  Y. Wu and
                  Daya Guo},
  title        = {DeepSeekMath: Pushing the Limits of Mathematical Reasoning in Open
                  Language Models},
  journal      = {arXiv preprint arXiv:2402.03300},
  year         = {2024},
  url          = {https://doi.org/10.48550/arXiv.2402.03300},
}

@article{Prasad2024selfconsistency,
  author       = {Archiki Prasad and
                  Weizhe Yuan and
                  Richard Yuanzhe Pang and
                  Jing Xu and
                  Maryam Fazel{-}Zarandi and
                  Mohit Bansal and
                  Sainbayar Sukhbaatar and
                  Jason Weston and
                  Jane A. Yu},
  title        = {Self-Consistency Preference Optimization},
  journal      = {arXiv preprint arXiv:2411.04109},
  year         = {2024},
  url          = {https://doi.org/10.48550/arXiv.2411.04109},
}

@inproceedings{babilong,
  author       = {Yuri Kuratov and
                  Aydar Bulatov and
                  Petr Anokhin and
                  Ivan Rodkin and
                  Dmitry Sorokin and
                  Artyom Y. Sorokin and
                  Mikhail Burtsev},
  title        = {BABILong: Testing the Limits of LLMs with Long Context Reasoning-in-a-Haystack},
  booktitle    = neurips,
  year         = {2024},
}

@inproceedings{chen-etal-2025-magicore,
    title = "{MA}g{IC}o{R}e: Multi-Agent, Iterative, Coarse-to-Fine Refinement for Reasoning",
    author = "Chen, Justin  and
      Prasad, Archiki  and
      Saha, Swarnadeep  and
      Stengel-Eskin, Elias  and
      Bansal, Mohit",
    booktitle = emnlp,
    month = nov,
    year = "2025",
}

@inproceedings{Wei2025cot,
  author       = {Jason Wei and
                  Xuezhi Wang and
                  Dale Schuurmans and
                  Maarten Bosma and
                  Brian Ichter and
                  Fei Xia and
                  Ed H. Chi and
                  Quoc V. Le and
                  Denny Zhou},
  title        = {Chain-of-Thought Prompting Elicits Reasoning in Large Language Models},
  booktitle    = neurips,
  year         = {2022},
}

@article{bachman2016informationseekingagents,
      title={Towards Information-Seeking Agents}, 
      author={Philip Bachman and Alessandro Sordoni and Adam Trischler},
      journal      = {arXiv preprint arXiv:1612.02605},
      year={2016},
      url={https://arxiv.org/abs/1612.02605}, 
}

@inproceedings{yuan202interactiveagent,
    title = "Interactive Machine Comprehension with Information Seeking Agents",
    author = "Yuan, Xingdi  and
      Fu, Jie  and
      C{\^o}t{\'e}, Marc-Alexandre  and
      Tay, Yi  and
      Pal, Chris  and
      Trischler, Adam",
    booktitle = acl,
    year = "2020",

}

\clearpage
\appendix
\crefalias{section}{appendix}

\section{Details of Experimental Setups \label{appendix:Implementation Details}}

\paragraph{Tool Use.}
We conduct all experiments within the Inspect-Eval evaluation framework~\citep{inspect_eval}, which fully supports the ReAct paradigm~\citep{Yao2023react} for multi-turn reasoning and tool interactions necessary for complex information-seeking tasks. The framework provides access to a comprehensive set of tools:
\begin{itemize}
    \item \textbf{Web Search}: We utilize the Serper search API to retrieve up-to-date web content for search queries.
    \item \textbf{Web Browsing}: The framework includes built-in browser automation tools, supporting essential web interaction functions such as browsing with URLs, clicking, scrolling down/up, typing, etc.
    \item \textbf{Code Execution}: The framework also supports built-in Python and Bash code execution environments.
\end{itemize}
In this evaluation framework, agents use these tools to interact with external sources to search and synthesize information to solve given information-seeking questions.  

\paragraph{Evaluation Benchmarks.}
We provide in-depth explanations of the long-horizon information-seeking benchmarks used in our experiments. 
(1) GAIA~\citep{Mialon2024gaia} evaluates the ability to act as a general AI assistant on complex retrieval and reasoning tasks spanning three difficulty levels.
Following prior work~\citep{Li2025webthinker, Gao2025beyondtenturns, kang2025acon}, we use 103 questions from the text-only validation subset. 
(2) WebWalkerQA~\citep{Wu2025webwalker} focuses on web-based reasoning, requiring agents to traverse webpages to locate target information across three difficulty levels.
We evaluate on 247 English questions. 
(3) FRAMES~\citep{Krishna2025frames} provides factual and reasoning-intensive queries to assess both retrieval and reasoning capabilities. We use a subset that consists of 300 samples that are randomly selected from the original dataset.

\paragraph{Baselines.}
In this section, we provide a more detailed explanation of the baselines.
\begin{itemize}[itemsep=2mm, parsep=1pt, leftmargin=*]
\item \textbf{GenPRM-7B}~\citep{Zhao2025genprm} is a generative PRM originally designed for mathematical reasoning. It produces Chain-of-Thought rationales with a binary verdict (\texttt{yes} / \texttt{no}) indicating whether the current step is correct.
We follow their prompt format and ask GenPRM to verify the correctness of a trajectory step and explain why the step is judged correct or incorrect.
\item \textbf{Web-Shepherd-8B}~\citep{Chae2025webshepherd} generates a task-specific checklist that decomposes a task into key subgoals and evaluates agentic trajectories based on it.
Specifically, it assigns coarse feedback labels (\texttt{Yes} / \texttt{No}/ \texttt{In progress}) for each checklist item for evaluation.
We also follow their prompt formats to generate a checklist for a given information-seeking task and evaluate each trajectory step relative to that checklist.
\item \textbf{StepWiser}~\citep{Xiong2025stepwiser} trains a generative PRM using GRPO with binary rewards, where each step is labeled as effective or ineffective.
For implementation, we follow the \textit{Relative Effective Reward Thresholding} in the paper to re-annotate our training dataset: a step receives a positive label if the ratio between the current and previous mean accuracies exceeds the threshold (0.7), and a negative label otherwise.
Using this binary supervision, we train Qwen3-4B for 4 epochs to build StepWiser PRM.
\item \textbf{Confidence}~\citep{Ghasemabadi2025guidedbygut, Fu2025deepthinkwconfidence, Prabhudesai2025maximizeconfidence, Wang2025eat}
estimates reasoning quality based on the model's certainty.
Following the confidence definition in recent work~\citep{Ghasemabadi2025guidedbygut, Fu2025deepthinkwconfidence}, we calculate confidence by taking the negative average log-probability of the top-10 tokens across all generated token positions in the reasoning step.
Higher scores indicate lower uncertainty.
\item \textbf{Relevance}~\citep{Wan2025relevance} measures the coherence between the current step and the preceding context. Specifically, it uses the Jaccard similarity between the current step and the accumulated past steps. A higher similarity indicates better contextual coherence.
\item \textbf{Verbal progress} is a zero-shot baseline that assesses progress toward the final answer by prompting Qwen3-4B to estimate how close the current reasoning state is to completing the task. The model is asked to output a scalar that ranges from 1 to 5 based on the textual content of the current step and its information-seeking trajectory. 
A higher score indicates that the current reasoning state is close to the final answer.
\end{itemize}

\begin{figure}[t]
    \centering
    \includegraphics[width=\linewidth]{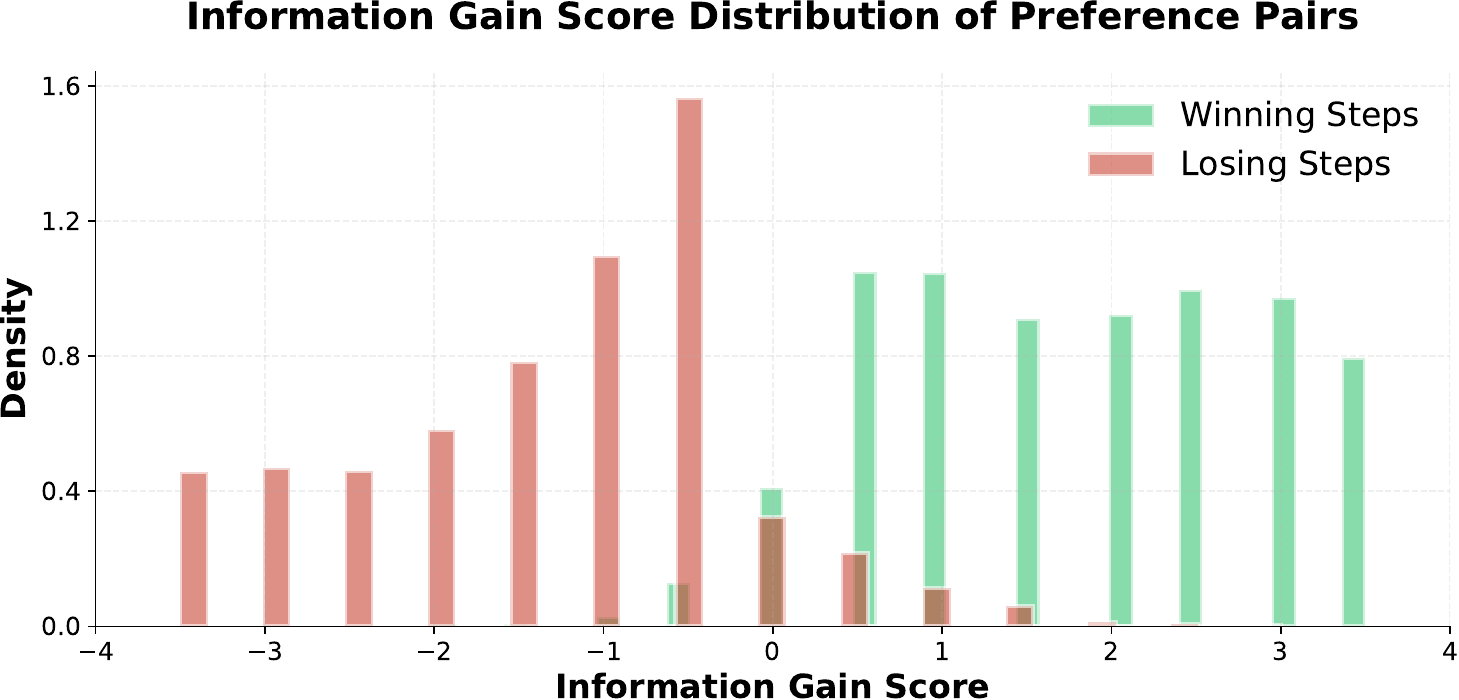}
    \par
    \caption{Distribution of annotated information gain scores. 
    }
    \label{fig:score_distribution}
\end{figure}

\paragraph{Train Configurations.}
We train \ours using Qwen3-4B with an alternating SFT-GRPO schedule over four cycles, where each cycle consists of one SFT epoch for summarization followed by one GRPO epoch for scoring, to jointly acquire both abilities.
A single SFT epoch requires approximately 0.5 hours and a single GRPO epoch requires 3.5 hours, totaling 16 hours on a single node with 8x NVIDIA RTX 6000 Ada GPUs (48GB).
For the SFT stage, we use a batch size of 128 and a learning rate of 1e-6.
For the GRPO stage, we execute $N=4$ rollouts and use a batch size of 128 and a learning rate of 1e-6. 
Such alternating optimization allows the model to continuously refine its summarization ability while simultaneously improving its scoring accuracy on reasoning quality. 
This iterative schedule ensures both modules evolve synergistically, stabilizing training and preventing either skill from degrading over time.

\paragraph{Data Annotations.}
We construct our annotated data from 4,344 information-seeking questions, comprising 720 questions used to train information-seeking agents from the Alibaba group~\citep{Wu2025webdancer, Li2025websailor, Tao2025webshaper}, and 3,624 questions from the Miroverse-v0.1 dataset~\citep{miromind2024opendata}, a large-scale agent dataset covering multi-hop QA, web navigation, and scientific reasoning tasks.
For the score annotation process, we execute $M=8$ rollouts per trajectory step to estimate the mean accuracy and information gain score. 
We discard steps that are either too easy ($m_t=1$) or too hard ($m_t=0$), resulting in 2,294 preference pairs used for training.

As shown in~\Cref{fig:score_distribution}, our annotation pipeline produces well-balanced information gain score distributions across the full score range.
This balanced distribution ensures that \ours learns to estimate diverse trajectory step quality, from harmful steps that reduce success probability to highly effective steps that substantially advance toward the correct answer.
Moreover, the distributions of winning and losing steps exhibit clear separation.
This clear separation validates the effectiveness of our preference pair construction and using this as training signals for \ours to output dense and comparative scores.

\paragraph{Frontier Model Performance.}
For the performance of frontier models on the GAIA benchmark, we follow the reported results (OpenAI DeepResearch: 67.4\% and DeepSeek-V3.1-671B: 63.1\%) in the DeepResearch-30B-A3B paper~\citep{tongyi2025deepresearch} and use these reported numbers as reference points when evaluating improvements brought by integrating \ours into DeepResearch-30B-A3B.

\section{Additional Experiments \label{appendix:Additional Experiments}}

\paragraph{\ours as Summarizer.}
To validate our design of jointly training scoring and summarization within a single model, we compare \ours against a variant that uses separate models for each capability.
Specifically, we train a PRM using only GRPO for scoring without SFT for summarization, and pair it with Qwen3-32B -- the same model we use for summary annotation -- to generate summaries at test-time.
As shown in~\ref{tab:summary_ablation}, \ours, which is jointly trained as both scorer and summarizer through our alternating SFT-GRPO schedule, outperforms this separated design.
This demonstrates that the two abilities are complementary and that our alternating training schedule enables seamless integration of these abilities.
We hypothesize that this benefit arises from positive transfer between the two objectives.
As both abilities operate on the same input (i.e., query, preceding summary, latest tool response, current trajectory step), learning to distill essential information during SFT directly aids GRPO optimization by highlighting the most relevant factors for quality evaluation.

\begin{table}[t]
    \centering
    \begin{minipage}{0.48\textwidth}
    \centering
    \small
        \renewcommand{\arraystretch}{1.0}
        \setlength{\tabcolsep}{6pt}
        \begin{tabular}{l c c c c}
             \toprule
             {\textbf{Method}} & {\textbf{FRAMES}} & \multicolumn{2}{c}{\textbf{GAIA}} & {\textbf{Avg.}}\\
              & & {\textbf{Level 1}} & {\textbf{Level 2}} & \\
             \midrule
             Qwen3-32B &
             {\small 54.7} & {\small 44.5} & {\small 29.5} & {\small 42.9} \\
             \ours &
             {\small \textbf{58.7}} & {\small \textbf{49.6}} & {\small \textbf{33.3}} & {\small \textbf{47.2}} \\
             \bottomrule
        \end{tabular}
    \caption{\textbf{Ablation study of summarizer.} The Qwen3-32B approach utilizes Qwen3-32B as a summarizer and employs the PRM trained solely as a scorer. \ours that simultaneously acts as a summarizer and scorer shows better performance, showing that two abilities are complementary.}
    \label{tab:summary_ablation}
    \end{minipage}
\end{table}

\begin{figure}[t]
    \centering
    \includegraphics[width=\linewidth]{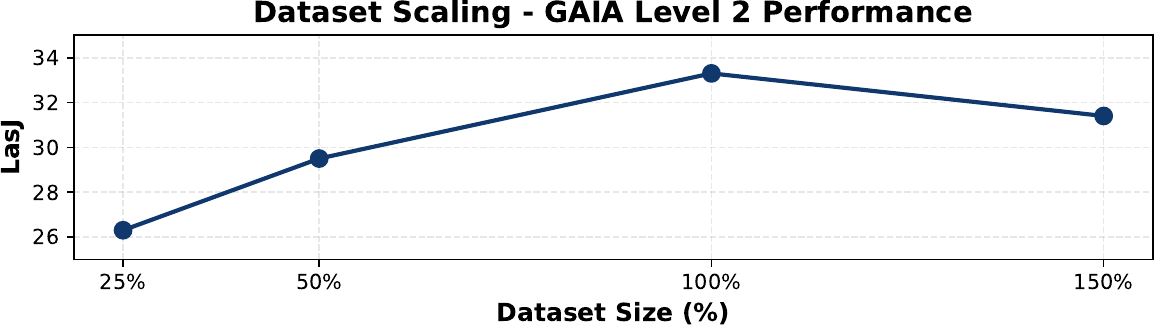}
    \par
    \caption{\textbf{Dataset scaling.} Experiments on the impact of dataset scaling on GAIA Level 2 using Qwen3-32B. Training \ours shows strong sample efficiency, achieving performance gain using only 50\%($\sim$1k samples) of our annotation data.}
    \label{fig:sub:dataset_scale}
\end{figure}
\begin{figure*}[t]
    \centering
    \includegraphics[width=\linewidth]{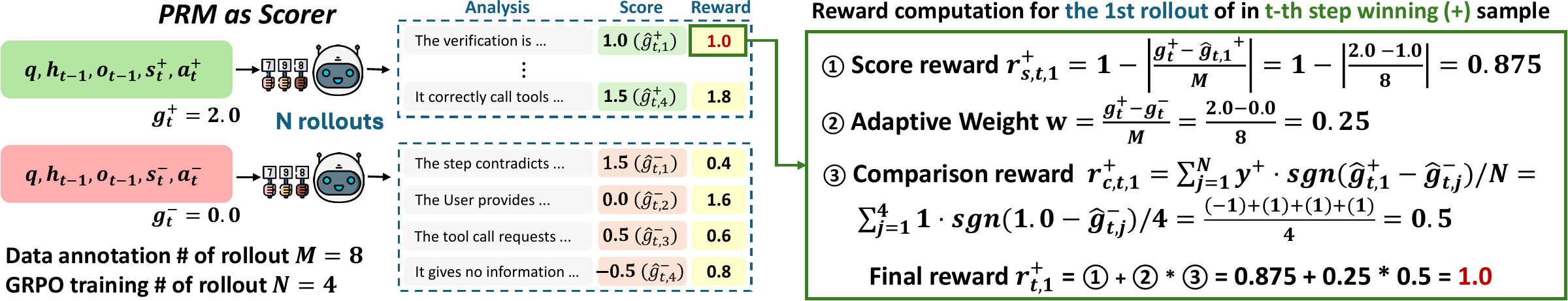}
    \par
    \caption{\small Example of reward computation for the first rollout of the t-th step winning sample during GRPO training.
    }
    \label{fig:reward_compute}
\end{figure*}

\paragraph{Dataset Scaling}
To further validate that \ours is a cost-efficient way of improving information-seeking behaviors of agents without fine-tuning them, we conduct an ablation study on the training dataset scaling.
As shown in~\Cref{fig:sub:dataset_scale}, using only 50$\%$ (i.e., $\sim$ 1k samples) of our annotation data still achieves 29.5 accuracy in GAIA Level 2, surpassing the base agent approach by training a relatively lightweight model (4B). 
The scaling curve saturates beyond 100\%, indicating that our $\sim$2k preference pairs represent a data-efficient point.
This demonstrates strong sample efficiency compared to fine-tuning agents, which typically require 10k-100k samples and expensive long-horizon rollouts using relatively larger models, requiring substantial computational resources.

\paragraph{Stability of Information Gain Estimation.}
As information gain in our pipeline is estimated via Monte Carlo rollouts, we conduct an empirical study to quantify the stochastic noise of this annotation process.
Using a subset of 400 query samples from our source dataset, we measure the mean accuracy (\Cref{eq:mean_acc}) for each trajectory step.
Treating a larger rollout budget ($M=32$) as a pseudo-ground truth reference, we bootstrap smaller subsets ($M \in \{4, 8, 16\}$ with $k=1000$ resamples to quantify estimation variance.
We observe that the variance of the estimated mean accuracy decreases significantly as the rollout count increases: 0.025 at $M=4$, 0.011 at $M=8$ and 0.004 at $M=16$.
Similarly, the Mean Absolute Error (MAE) relative to the $M=32$ reference drops from 0.097 to 0.037.
To evaluate the downstream impact on preference learning, we calculate the preference flip rate --- how often rollout noise alters the preferred step relative to the pseudo-ground truth.
This rate is notably low, decreasing from 3.62\% ($M=4$) to 1.81\% ($M=16$). 
These results indicate that the quality differences between winning and losing steps typically exceed Monte Carlo noise, confirming that our supervision signal reflects robust quality distinctions rather than sampling variance.

\paragraph{Impact of Annotation Policy Model.}
We further analyze the effect of the rollout policy model's capability on the annotation pipeline by comparing Qwen3-32B and Qwen3-14B on the same 400-query subset ($M=8$). 
Both models yield consistent average information gain scores (\Cref{eq:information_gain}) for winning and losing samples (1.87 and -1.21 for Qwen3-32B; 1.73 and -1.16 for Qwen3-14B), demonstrating that our pipeline generates stable preference signals across model scales. 
However, because our annotation process terminates when a step reaches 100\% or 0\% success, a sufficiently capable model is required to maximize the learning signal. 
The 32B model maintains a non-trivial success probability over more steps, thereby avoiding premature termination. 
Consequently, Qwen3-32B generates a higher yield of preference pairs (240 vs.\ 169) and exhibits greater score diversity, where Qwen3-32B results in standard deviations of 1.27 and 1.30 for winning and losing samples, compared to 1.20 and 1.03 for the 14B model. 
This better alignment between policy capability and query difficulty justifies our selection of Qwen3-32B as the optimal annotator.

\paragraph{Qualitative Analysis of Failure Cases.}
To better understand \ours, we inspected information-seeking trajectories from challenging subsets (i.e., GAIA Levels 2 and 3) generated by the Qwen3-32B agent and \ours.
Our analysis reveals that \ours occasionally exhibits a failure mode of over-conservatism.
Qualitatively, the model frequently assigns negative informativeness scores to final answers if the agent arrives at them through speculation or with unresolved uncertainties, rather than relying on explicit, verifiable evidence.

While this conservative behavior is generally beneficial for preventing the agent from concluding with unverified answers, it can inadvertently force the agent to discard a correct heuristic shortcut.
Consequently, the agent is compelled to continue searching to clear any remaining uncertainties, which prolongs the trajectory, thereby increasing the risk of context overflow or step-limit timeouts.
For instance, in a representative GAIA Level 3 trajectory, the agent correctly guesses the target answer "mice" but is penalized by \ours for lacking definitive textual evidence.
As a result, the model favors continuing finding concrete evidence over terminating the trajectory.
This shows a fundamental trade-off in our approach between strict factual grounding and trajectory efficiency.

\begin{figure*}[t]
    \centering
    \includegraphics[width=\linewidth]{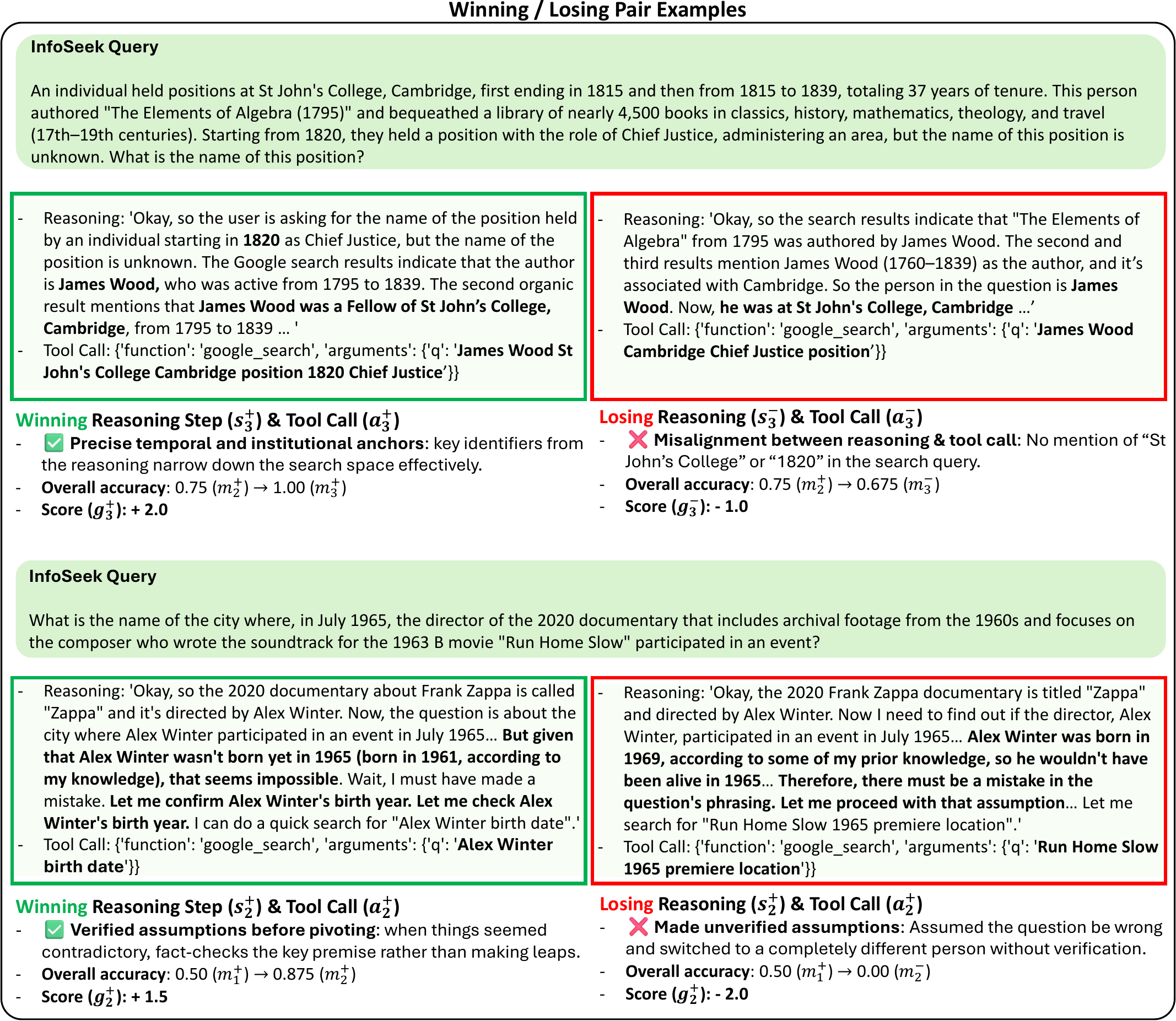}
    \par
    \caption{\small Examples of preference pairs constructed by our annotation pipeline.
    }
    \label{fig:win_lose_examples}
\end{figure*}
\begin{figure*}[t]
    \centering
    \includegraphics[width=\linewidth]{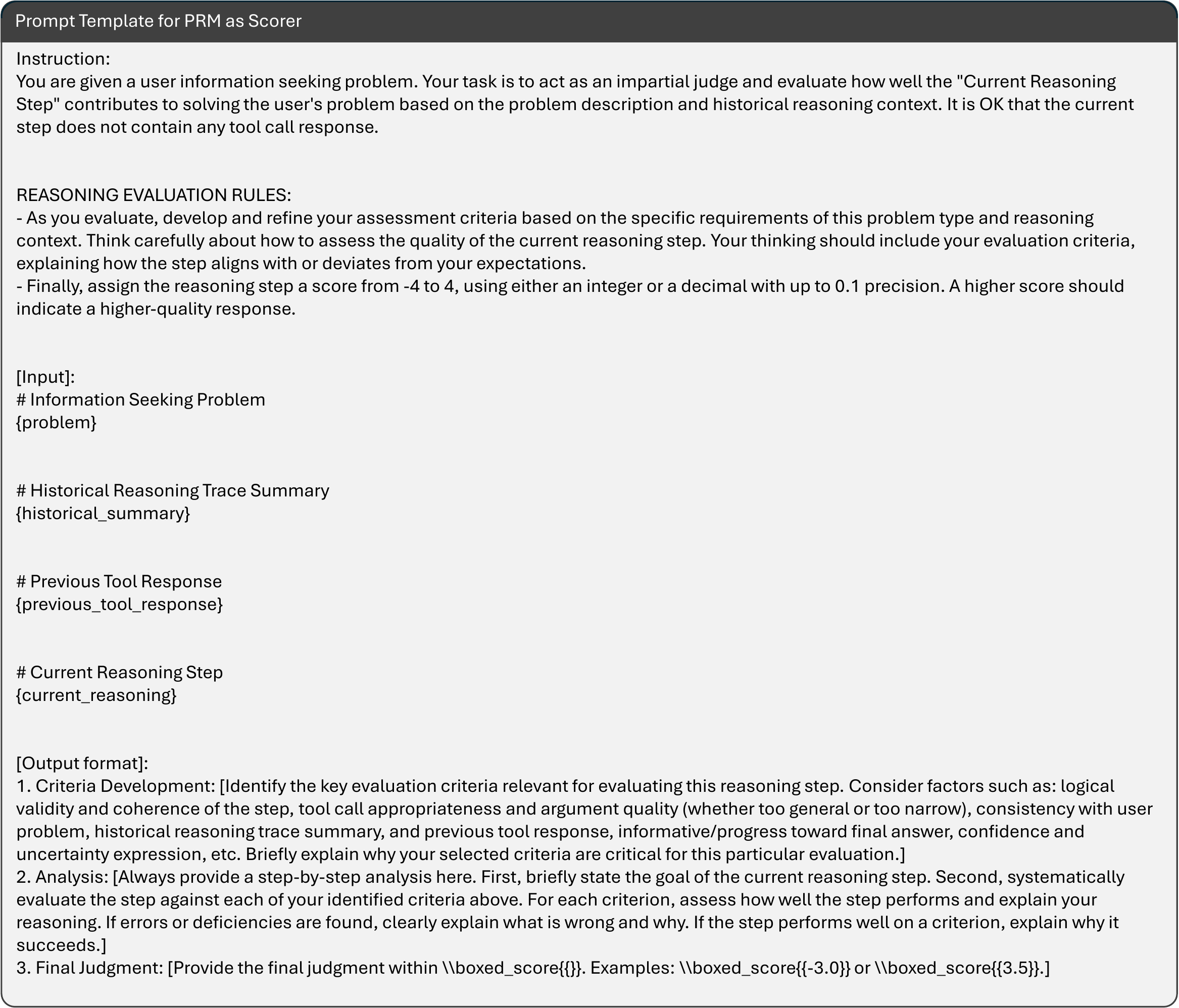}
    \par
    \caption{\small Input prompt for \ours when the model is trained with GRPO for scoring ability and acts as a scorer at test-time.
    }
    \label{fig:sub:prompt_score}
\end{figure*}
\begin{figure*}[t]
    \centering
    \includegraphics[width=\linewidth]{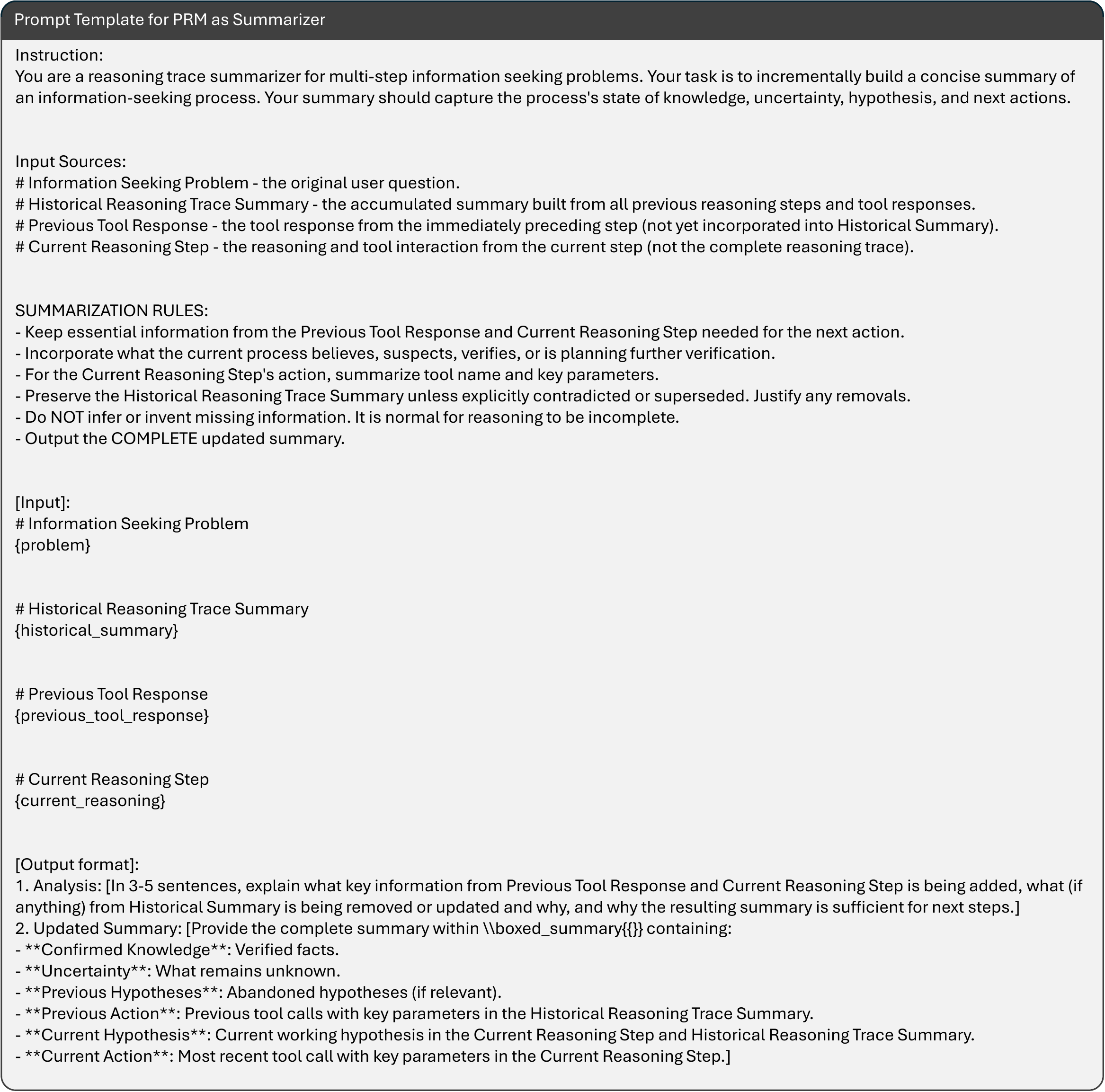}
    \par
    \caption{\small Input prompt for \ours when the model is trained with SFT for summarization ability and acts as a summarizer at test-time.
    }
    \label{fig:sub:prompt_summary}
\end{figure*}
\begin{figure*}[t]
    \centering
    \includegraphics[width=\linewidth]{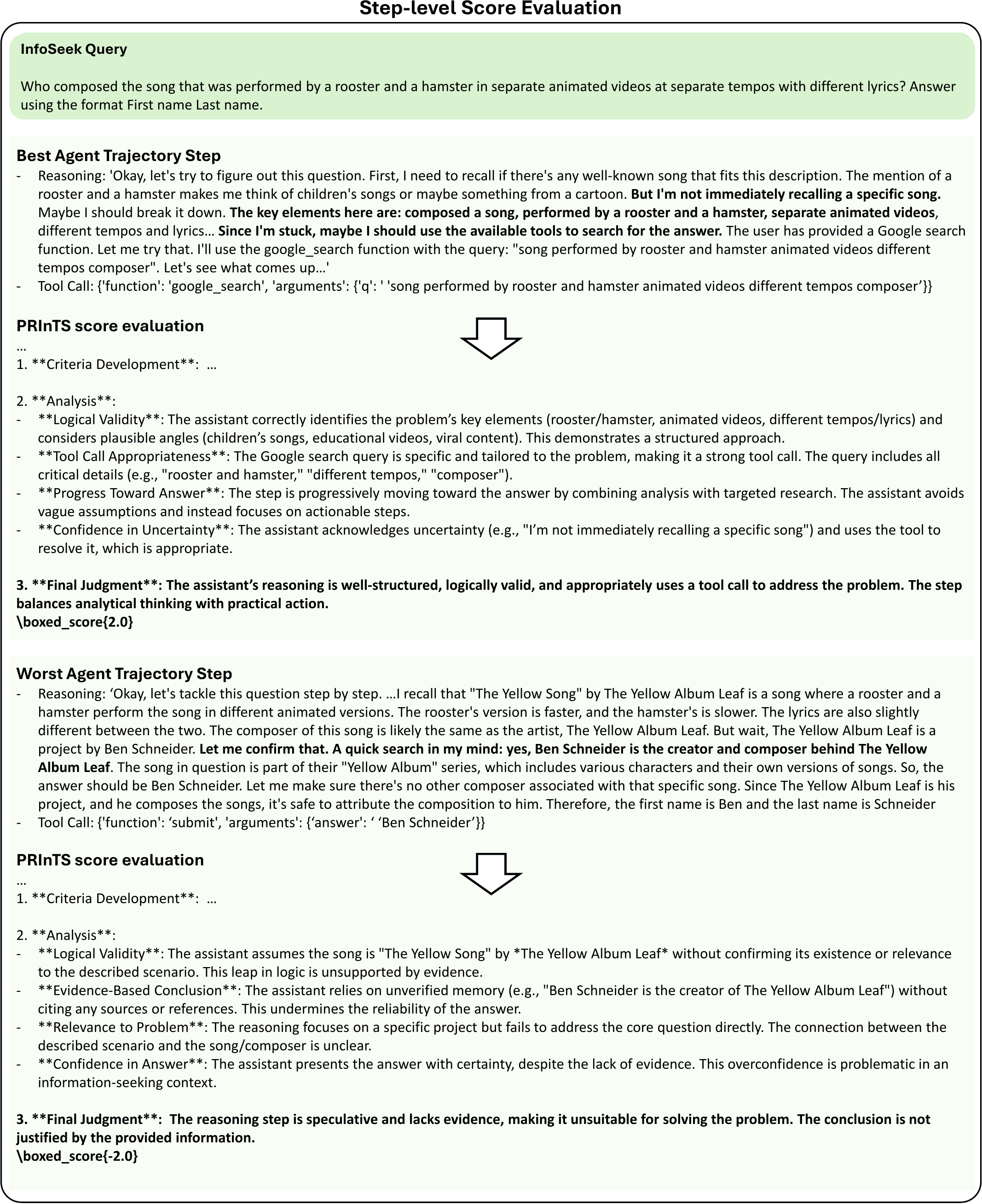}
    \par
    \caption{\small \ours step-level evaluation examples on a GAIA query. Among four candidate steps, we show the highest-scoring (top) and lowest-scoring (bottom) steps. The high-quality step acknowledges uncertainty and initiates an appropriate tool call to gather missing information, while the low-quality step makes unverified assumptions and confidently produces an unsupported answer without evidence.
    }
    \label{fig:prm_output}
\end{figure*}

\end{document}